\documentclass{article}

\PassOptionsToPackage{numbers, compress}{natbib}

\usepackage[eandd, final]{neurips_2026}

\usepackage[utf8]{inputenc} 
\usepackage[T1]{fontenc}    
\usepackage{hyperref}       
\usepackage{url}            
\usepackage{booktabs}       
\usepackage{amsfonts}       
\usepackage{nicefrac}       
\usepackage{microtype}      
\usepackage{xcolor}         
\usepackage[table]{xcolor} 
\usepackage{adjustbox} 
\usepackage{graphicx}
\usepackage{multirow} 
\usepackage{booktabs} 
\usepackage{pifont} 
\usepackage{amsmath}
\usepackage{longtable}
\usepackage{array}

\title{MotionHalluc: Diagnosing Kinematic Hallucinations in Fine-Grained Motion Reasoning}

\author{%
    Weile Guo\textsuperscript{1},
    Shenghong He\textsuperscript{1},
    Danying Mo\textsuperscript{1},
    Chengdong Xu\textsuperscript{1},
    Xuexun Liu\textsuperscript{2},
    Chao Yu\textsuperscript{1}\thanks{Corresponding author.} \\
    \textsuperscript{1}Sun Yat-Sen University \\
    \textsuperscript{2}Shenzhen University \\
    \texttt{kwokwlok@mail2.sysu.edu.cn}
}

\begin{document}

\maketitle

\begin{abstract}
  Motion instruction generation in cross-video comparison aims to produce corrective feedback that describes the differences between a query and a reference motion. However, existing models often generate instructions that exhibit motion hallucinations, failing to reflect actual kinematic differences between paired videos. To systematically investigate these hallucinations, we introduce MotionHalluc, a dedicated benchmark for evaluating motion hallucinations in paired-video comparison. MotionHalluc comprises 1540 fine-grained questions over 553 video pairs, evaluating hallucinations along three core dimensions: (1) \textbf{directional} hallucination, (2) \textbf{attributional} hallucination, and (3) \textbf{temporal} hallucination. Extensive evaluations of state-of-the-art large multimodal models demonstrate high susceptibility to these hallucinations. Furthermore, we provide Perceive-Parse-Verify (PPV) as a training-free measurement extraction and verification baseline that converts candidate instructions into executable measurement queries and supplies kinematic measurements at inference time. Our results show that this simple measurements injection yields an average 10.6\% performance gain across models, suggesting that motion reasoning with explicit quantitative measurements is a key factor in reducing hallucinations in cross-video comparison. Our code and dataset are available at \url{https://github.com/Leocoding0971/MotionHalluc}.
\end{abstract}

\section{Introduction}

Recent advances in large multimodal models (LMMs) have significantly improved visual understanding \cite{hurst2024gpt, an2025llava, wang2025internvl3_5, Qwen2.5-VL, comanici2025gemini}, enabling progress on a wide range of video-based tasks. Among them, motion instruction generation \cite{fang2024cigtime} has emerged as an important setting for fine-grained motion understanding \cite{yeh2025coachme, li2025unipose, delmas2023posefix, du2025motionsight}. In this task, given a query video and a reference video, the model is required to generate corrective instructions that describe how the query motion should be adjusted to better match the reference. This task is particularly valuable in applications such as skill coaching, exercise assessment, and rehabilitation feedback \cite{parmar2022domain, anand2022yoga, verma2020yoga, wang2026integrating}, where precise and interpretable motion guidance is essential. 

However, in practice, we observe that LMMs frequently generate instructions that are linguistically plausible but inconsistent \cite{bae2025mash} with the actual motion differences between the two videos. Even when the query and reference motions are highly similar, models tend to produce spurious instructions driven by visual biases such as body shape, viewpoint, or appearance, rather than actual motion differences \cite{ye2026mm, han2024instinctive}. These issues stem from a broader limitation in prior work \cite{kong2025mhbench, nie2024slowfocus, li2025vidhalluc}, which lacks explicit modeling of cross-video relational reasoning, fine-grained spatial grounding, and temporally consistent comparison.

\begin{figure} 
    \centering
    \includegraphics[width=\linewidth, trim=0mm 40mm 0mm 0mm, clip]{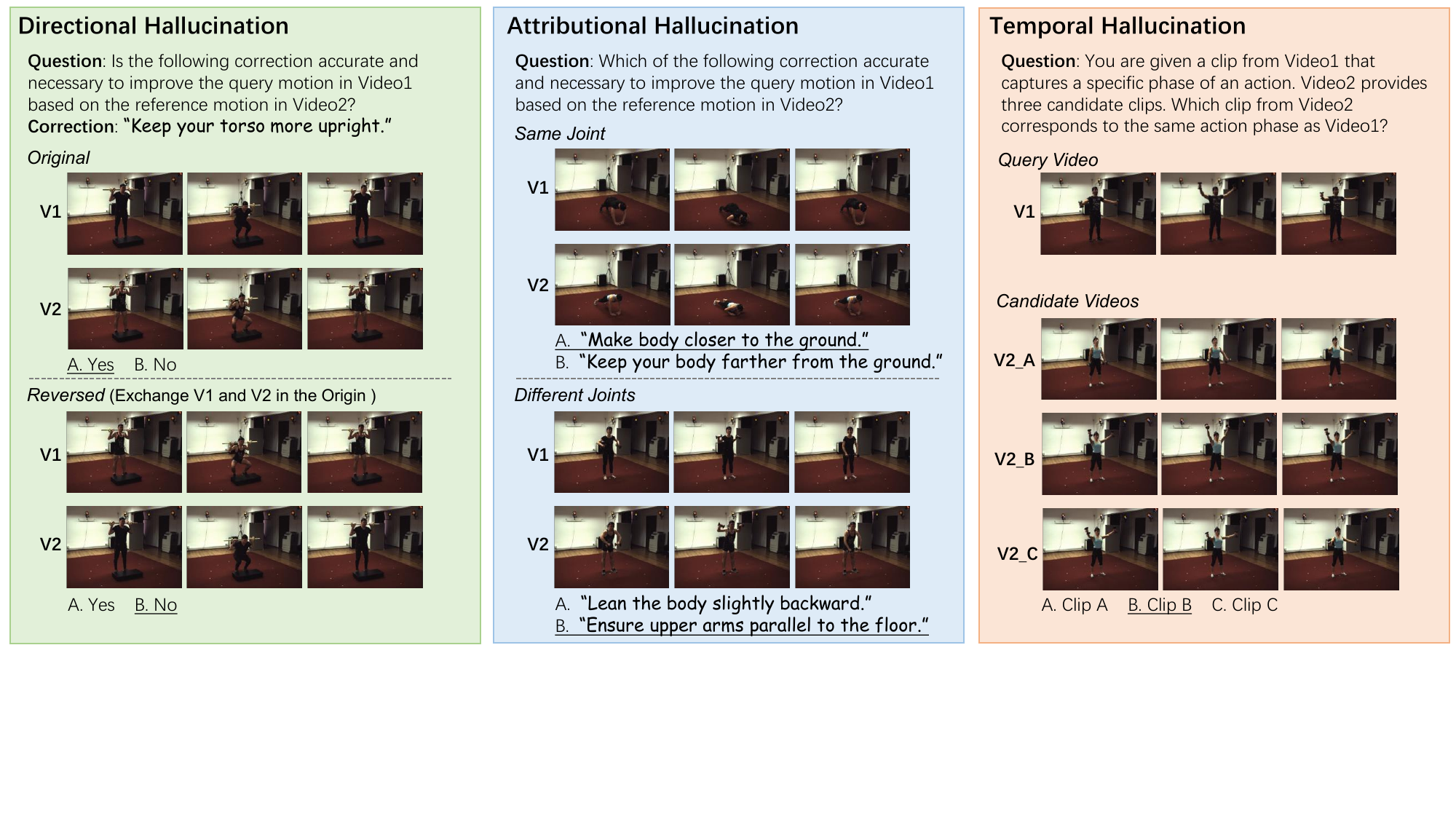} 
    \caption{Examples of three hallucination types in MotionHalluc. Given paired videos (V1 as query, V2 as reference), the model selects an answer from candidates. \textbf{Directional Hallucination} (left): determine whether a direction-related instruction is accurate and necessary, with an additional reversed query–reference setting. \textbf{Attributional Hallucination} (middle): choose the instruction that refers to the correct body attribute or joint. \textbf{Temporal Hallucination} (right): select the clip that matches the same action phase as the query. The correct answers are \underline{underlined}.}
    \label{fig:examples} 
\end{figure}

To systematically study these problems, we introduce MotionHalluc, the first benchmark specifically designed to evaluate kinematic hallucinations in fine-grained motion reasoning over cross-video comparison. We design a set of controlled evaluation protocols that explicitly test whether an instruction is consistent with the motion differences between the query and reference videos. Specifically, MotionHalluc evaluates hallucinations along three critical dimensions (\autoref{fig:examples}) : (1) Directional Hallucination (DH), which tests whether the model understands the logical direction of an instruction by distinguishing the reference from the query; (2) Attributional Hallucination (AH), which assesses if the model can accurately pinpoint the specific body part or movement error among multiple options; and (3) Temporal Hallucination (TH), which determines if the model is correctly aligning and comparing the same phases of an action across different clips. These evaluation settings allow us to quantify different failure modes in a precise and interpretable manner, going beyond existing general video hallucination and fine-grained motion benchmarks \cite{li2025videohallu, kong2025mhbench}, which often lack cross-video comparison.

Extensive experiments on MotionHalluc reveal that state-of-the-art LMMs are highly susceptible to all three types of hallucinations. Although the paired-video setup provides a strong inductive bias for comparison, it is insufficient to ensure that models reason over actual motion differences rather than superficial cues or prior biases. To this end, we provide Perceive-Parse-Verify (PPV) as a training-free kinematic measurement extraction and verification baseline for MotionHalluc: PPV firstly reconstructs motion from videos, then translates candidate instructions into executable measurement queries, and outputs evidence that can be used to support (or refute) a model's decision at inference time.

In summary, our contributions are threefold:

\begin{itemize}

\item We introduce MotionHalluc, a novel benchmark for studying motion hallucinations in fine-grained motion reasoning over cross-video comparison, targeting three key hallucinations (Directional Hallucination, Attributional Hallucination, and Temporal Hallucination), and integrating atomic corrective instructions with 3D motion sequences to systematically evaluate this unexplored problem in paired-video motion reasoning.

\item We conduct a systematic evaluation of five state-of-the-art LMMs under these controlled protocols and provide an in-depth analysis of model failure patterns and biases, revealing that paired-video inputs alone are insufficient for reliable kinematic comparison.

\item We provide PPV as a training-free kinematic measurement extraction and verification baseline that produces quantitative evidence to support verifiable judgments, and show that such evidence can substantially reduce hallucinations on MotionHalluc, suggesting that explicit kinematic measurement is beneficial for improving the reliability of multimodal reasoning systems.

\end{itemize}

\section{Related work}
\label{related_work}

\paragraph{Motion Understanding and Instruction Generation.} Video motion understanding and instruction generation is a long-term research in the field of computer vision. Traditional Action Quality Assessment (AQA) primarily focuses on predicting scalar scores to measure performance quality \cite{parmar2019and, pirsiavash2014assessing, dong2024interpretable}. While these methods have achieved significant progress in domains such as diving and gymnastics, they are unable to provide fine-grained linguistic feedback to improve user performance \cite{henriques2026can, wang2025attention}. To overcome this limitation, motion instruction generation has been introduced, which shifts the setting to paired-video scenarios where a given execution is explicitly compared against a reference performance. Existing motion instruction approaches, such as CoachMe \cite{yeh2025coachme}, attempt to generate personalized coaching suggestions through reference-based decoding, while domain-specific studies in yoga and fitness utilize skeleton-based visualizations to facilitate error correction \cite{wu2021computer, li20223d, tharatipyakul2024deep}.

\paragraph{Hallucination Evaluation in LMMs.} Despite the impressive perceptual capabilities of state-of-the-art LMMs, their reliability in complex motion reasoning remains unexplored. Existing hallucination research has primarily targeted object existence, counting, or coarse action recognition in static images or general video tasks \cite{li2023evaluating, li2025vidhalluc, luo2025dr, wu2025mass}. Systematic definitions and quantifications of kinematic-dimensional hallucinations remain absent in the context of paired-video comparison. Furthermore, traditional motion-comparison datasets rely heavily on linguistic metrics like BLEU \cite{papineni2002bleu} or ROUGE \cite{lin2004rouge} to evaluate the text similarity of expert comments and the generated instructions. However, these metrics fail to capture the alignment between instructions and underlying kinematic facts (e.g. actual joint angles or positions) \cite{henriques2026can}. While LLM-based evaluators like G-Eval \cite{liu2023g} offer improved explainability, their logic is still rooted in linguistic priors rather than kinematic grounding \cite{gao2025exploring}. Recent work, such as VidDiff \cite{burgess2025video}, explores fine-grained video difference description; similarly, ConViS \cite{liberatori2025convis} focuses on evaluating video similarity through semantic concepts. Nevertheless, these benchmarks focus on models' comparison ability rather than diagnosing the structural hallucination flaws of LMMs. In contrast, our proposed MotionHalluc is the first benchmark specifically designed for the motion hallucination evaluation in cross-video comparison. MotionHalluc defines three core hallucination dimensions, providing a novel diagnostic tool for evaluating the kinematic reliability of LMMs.

\paragraph{Motion-Language Fusion Models.} Structured motion data, such as 3D skeletal sequences and SMPL parameters \cite{DBLP:journals/tog/LoperM0PB15}, are vital for action recognition \cite{chen2025skeleton, liu2022end} and motion estimation \cite{goel2023humans, shin2024wham}, highlighting the importance of kinematic grounding in multimodal learning. In the context of instruction generation, several works have integrated motion features with visual encoders. For instance, MotionGPT \cite{jiang2023motiongpt} and Motion-LLM \cite{chen2025motionllm} treat motion sequences as a foreign language by tokenizing motion data into the original embedding space to enhance perception. However, these fusion models \cite{zhang2024motiongpt, zhu2506motiongpt3, yeh2025coachme} often suffer from high data dependency on expensive motion capture datasets and exhibit limited generalization, often resulting in models that specialize in a narrow set of actions. More importantly, existing paradigms \cite{hu2025hmvlm, wu2024motion} mostly treat motion data as a means of perception enhancement, while overlooking its potential as evidence for logic verification. There is a notable lack of work that can translate natural language corrective instructions into executable kinematic queries and leverage motion data for instruction validation. To address this limitation, we propose PPV, which is a training-free pipeline that integrates perception, parsing, and verification. Unlike prior end-to-end training methods that heavily rely on large-scale paired motion-language data \cite{chen2025motionllm, li2025unipose, fang2024cigtime}, PPV introduces explicit kinematic measurements at inference time, bridging the gap between semantic reasoning and physical reality.

\section{MotionHalluc Benchmark}
\label{sec:benchmark}

We seek to advance motion understanding by introducing MotionHalluc, a benchmark specifically designed to evaluate motion hallucinations in cross-video comparison. Existing video-pair benchmarks are not optimized for diagnosing kinematic reasoning flaws: ConViS-Bench \cite{liberatori2025convis} focuses on concept-based similarity scoring, while VidDiff \cite{burgess2025video} centers on matching videos to textual descriptions. Although general hallucination benchmarks exist \cite{kong2025mhbench, li2025vidhalluc}, they lack the comparative settings required for fine-grained motion reasoning. We bridge this gap by refining motion evaluation through directional, attributional, and temporal dimensions to systematically identify hallucinations. Furthermore, MotionHalluc incorporates manual corrective instructions and aligned 3D motion data to provide broad scalability for future related work (\autoref{tab:dataset-comparison}).

\subsection{Data Collection}
\label{sec:data_collection}

\paragraph{Video and Motion Collection.}
The raw videos in MotionHalluc are collected from the Fit3D dataset \cite{fieraru2021aifit}, which provides synchronized multi-view recordings and accurate 3D motion-capture data for a wide range of fitness exercises. From the original set of 47 action categories, we select 32 actions with clear biomechanical structure, observable form variations, and sufficient inter-subject diversity. These properties are essential for constructing reliable motion comparisons and identifying meaningful differences. Each recording contains multiple repetitions of the same exercise by a single subject, which exhibit highly similar motion patterns and may introduce redundancy and bias. To address this, we retain the second repetition for each subject, as it avoids the preparation phase often present in the first repetition and better reflects steady execution. This reduces redundancy while preserving diversity across subjects. To enable kinematic verification, these recordings are aligned with 3D motion data, converted from SMPL-X \cite{pavlakos2019expressive} into a 22-joint skeleton following HumanML3D conventions \cite{Guo_2022_CVPR}. Since the video and motion-capture streams were synchronously recorded at 50 FPS, temporal alignment is naturally preserved through consistent segmentation, allowing for the direct extraction of kinematic measurements corresponding to specific video segments.

\begin{table}[t]
\centering
\caption{Comparison of MotionHalluc with existing video comparison datasets. Our benchmark is the first to explicitly target motion hallucination in cross-video comparison, featuring fine-grained annotations and aligned 3D ground truth data for kinematic verification.}
\label{tab:dataset-comparison}
\resizebox{\columnwidth}{!}{
\begin{tabular}{lccccccc}
\toprule
\textbf{Dataset} & \textbf{Pairs} & \textbf{Evaluation Objective} & \textbf{BinaryQA} & \textbf{MCQ} & \textbf{Instructions} & \textbf{Timestamp} & \textbf{3D Motion} \\ \midrule
VidDiffBench & 549 & Difference Discovery & \ding{51} & \ding{55} & \ding{55} & \ding{51} & \ding{55}\\ 

ConViS-Bench & 610 & Similarity Scoring & \ding{55} &\ding{55} & \ding{55} & \ding{55} & \ding{55} \\ 

MotionHalluc (Ours) & 553 & Hallucination Diagnosis & \ding{51} & \ding{51} & \ding{51} & \ding{51} & \ding{51} \\ 
\bottomrule
\end{tabular}
}
\end{table}

\paragraph{Video Pairing.}
Our benchmark focuses on analyzing fine-grained motion differences between paired videos. For each pair, one video is treated as the query and the other as the reference. To gather pairs, we first apply Dynamic Time Warping (DTW) \cite{sakoe1978dynamic} to temporally align randomly paired videos within the same action category. Based on the alignment results, we prune redundant head and tail segments (typically irrelevant preparation or idle phases) to isolate and synchronize the core motion execution. Following this, the Euclidean distance between aligned sequences is used as a coarse measure of motion similarity. To guarantee that the pairs exhibit visually discernible yet subtle differences, we exclude candidates with too low or too high similarity scores. Finally, all remaining pairs are manually inspected. This rigorous filtering process removes approximately 38\% of the initial candidates, resulting in a final set of 553 high-quality video pairs.

\paragraph{Human Annotation.}
For each query–reference video pair, annotators provide a list of short, imperative instructions describing the most salient motion differences. To ensure objectivity, these descriptions are strictly grounded on observable kinematic differences, such as joint angles, body-part positions, where speculative or non-observable factors are explicitly excluded. This process is supported by a custom-developed interface that presents synchronized multi-view videos for both query and reference, enabling precise, frame-level inspection. To further localize these motion differences, each corrective instruction is associated with a specific pair of frames (one from each video) that serves as visual evidence of the described discrepancy. These frame-level annotations provide a rigorous foundation for constructing evaluation tasks. Further details about the annotation process are provided in the \autoref{sec:human_annotation}.

\paragraph{Data Curation.}
On average, annotators spend approximately 3.9 minutes per video pair. To ensure high annotation quality, we adopt a two-stage verification. In the first stage, a primary annotator provides corrective instructions grounded in observable motion differences. In the second stage, a separate annotator performs a full review of all annotations, checking each instruction with visual observation and identifying ambiguous or potentially misleading descriptions. During this process, approximately 6.5\% of the initial annotations are identified as ambiguous and are subsequently revised and corrected. This rigorous procedure results in a cleaner and more reliable dataset, with an average of 4.8 detailed annotations per video pair (including corrective instructions and timestamps).

\begin{table}
\centering
\caption{MotionHalluc statistics.}
\label{tab:motionhalluc-statistics}
\resizebox{\columnwidth}{!}{
\begin{tabular}{lccccccc}
\toprule
\textbf{Dataset} & \textbf{Action Categories} & \textbf{Instructions} & \textbf{Timestamp} & \textbf{DH} & \textbf{AH} & \textbf{TH} & \textbf{Total QA} \\ \midrule
MotionHalluc (Ours) & 32 & 891 & 1782 & 624 & 600 & 316 & 1540 \\ 

\bottomrule
\end{tabular}
}
\end{table}

\subsection{Evaluation Protocol for Motion Hallucination}
\label{sec:evaluation_protocol}

To systematically diagnose motion hallucinations in paired-video settings, we design three different tasks. Each task targets a distinct failure mode: Directional Hallucination (DH), Attributional Hallucination (AH), and Temporal Hallucination (TH). All tasks are formulated as closed-form question answering problems, including binary QA and multiple choice questions, enabling precise and interpretable evaluation. \autoref{tab:motionhalluc-statistics} summarizes the statistics for MotionHalluc. For all tasks, we report accuracy as the evaluation metric.

\paragraph{Directional Hallucination.} 
DH is evaluated using triplets $(v_1, v_2, i)$, where $v_1$ is the query video, $v_2$ is the reference, and $i$ is the human-annotated instruction. We construct two highly confusable settings to diagnose this failure mode: \textbf{Original instances} follow the original sequence $(v_1, v_2, i)$, where $i$ accurately describes the transformation required for $v_1$ to match $v_2$. \textbf{Reversed instances} exchange the video order to $(v_2, v_1, i)$ while still designating the first video as the query, thereby breaking the directional consistency between the motion pair and the instruction $i$. The model is then asked to verify the validity of $i$ under both settings. This task isolates whether the model performs genuine relative reasoning or simply relies on the standalone semantic plausibility of the instruction text.

\paragraph{Attributional Hallucination.}
AH evaluates whether models accurately localize and attribute motion differences through a forced-choice task involving a video pair $(v_1, v_2)$ and two candidate instructions $(i_1, i_2)$. Similarly, two settings are devised to probe this failure mode. \textbf{Same Joint}: We select a ground-truth instruction $i_{gt}$ and employ GPT-5 \cite{singh2025openai} to generate a semantically opposite instruction for the same joint (e.g., "bend legs" $\to$ "straighten legs") to form the pair $(i_1, i_2)$. \textbf{Different Joints}: We sample two ground-truth instructions targeting different body parts for the same pair; we then randomly select one and perturb it into an incorrect variant via GPT-5, resulting in a pair $(i_1, i_2)$ where only one option is kinematically valid.

\paragraph{Temporal Hallucination.}
TH arises when a model fails to align corresponding motion phases across videos, leading to incorrect reasoning about when a movement occurs. To evaluate temporal alignment, we leverage frame-level annotations to construct clip-level matching problems. For each instance, we extract a short query clip $v_1^{clip}$ centered around an annotated frame, and sample multiple candidate clips $(v_2^A, v_2^B, v_2^C)$ from different temporal segments of the reference video. Only one candidate corresponds to the same motion phase as the query, while the others are drawn from distinct phases of the same action, making them visually similar but temporally inconsistent. The model is required to identify the temporally aligned segment. This setting emphasizes fine-grained temporal reasoning and exposes failures where models rely on coarse appearance similarity instead of precise motion dynamics.

\begin{figure} 
    \centering
    \includegraphics[width=\linewidth, trim=5mm 35mm 5mm 0mm, clip]{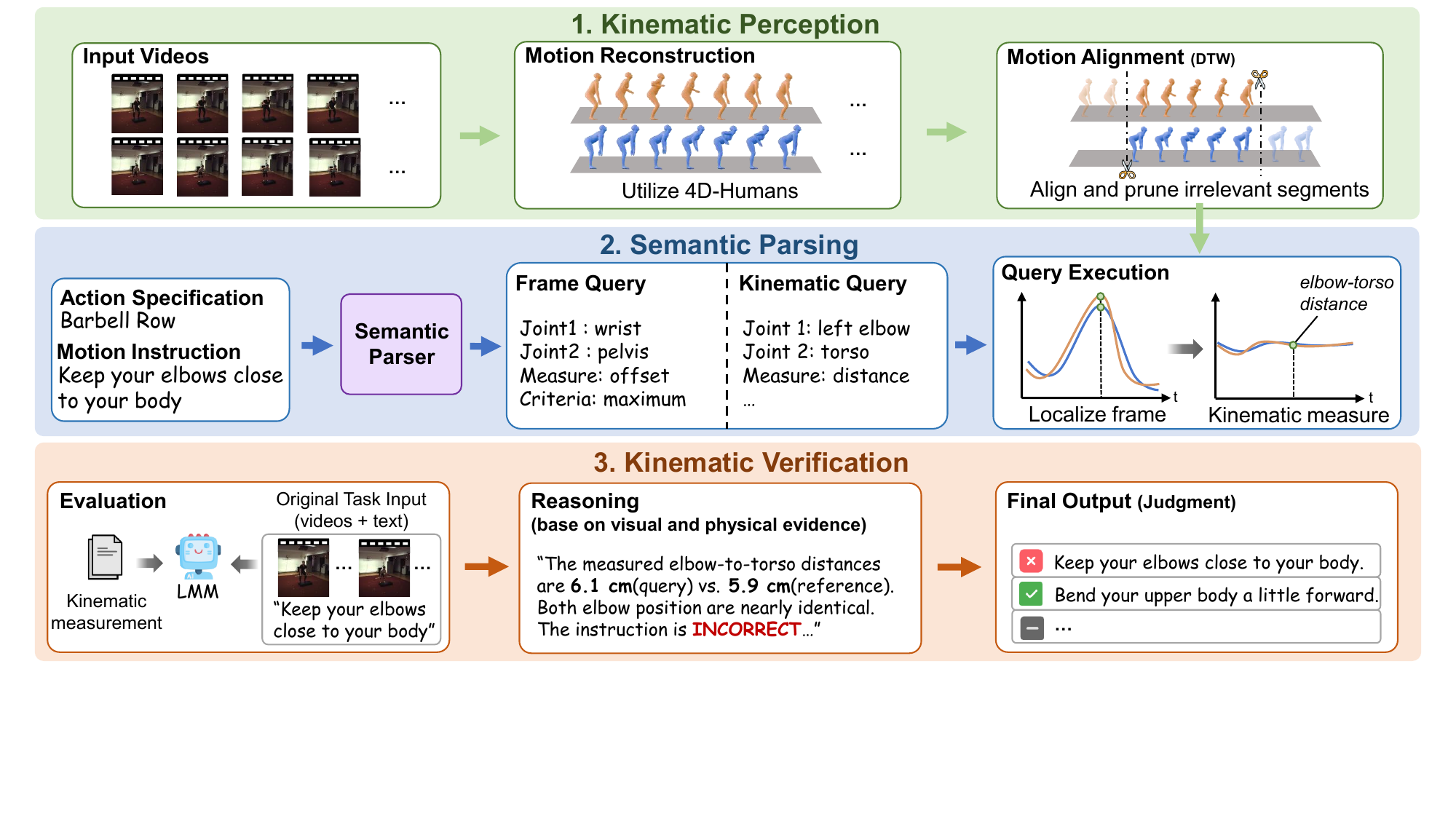} 
    \caption{The 3-stage PPV pipeline. \textbf{Kinematic Perception}: Reconstruct motion sequences from video pairs and align them. \textbf{Semantic Parsing}: Parse text into executable queries defining when and what to measure, and execute them to obtain kinematic measurements. \textbf{Kinematic Verification}: Reason over visual and kinematic measurements to produce a grounded judgment.}
    \label{fig:main_method} 
\end{figure}

\section{PPV Pipeline}
\label{sec:method}

We introduce Perceive-Parse-Verify (PPV) as a training-free, evidence-augmented verification pipeline (\autoref{fig:main_method}) that provides explicit kinematic measurements without modifying the evaluated model's learned visual representations. PPV consists of three stages: (1) \textbf{Kinematic Perception}, which reconstructs and aligns 3D skeletal sequences from raw videos to establish a physical basis; (2) \textbf{Semantic Parsing}, which translates natural language instructions into executable kinematic queries to identify key frames and measurements; and (3) \textbf{Kinematic Verification}, which augments the model's reasoning by providing it with extracted evidence for grounded judgment.

\paragraph{Kinematic Perception.}
Given the query and reference video inputs $(v_q, v_r)$, we first reconstruct motion sequences $(m_q, m_r)$ from input videos using 4D-Humans \cite{goel2023humans}. We then perform temporal alignment between motion sequences using DTW \cite{sakoe1978dynamic}, followed by the removal of irrelevant segments (e.g., preparation or idle phases), ensuring that subsequent analysis focuses on the core motion dynamics. This step transforms raw pixels into structured motion data, which establishes a physical basis for verifying the corrective instructions.

\paragraph{Semantic Parsing.}
To bridge the gap between natural corrective language and kinematic grounding, PPV converts the task inputs (text only) into structured, executable queries. Specifically, given a motion instruction $i$ and action specification $\mathcal{A}$, the semantic parser $\mathcal{P}$ (LMM) generates two different queries, $Q_f$ and $Q_k$:

\begin{equation}
\{Q_f, Q_k\} = \mathcal{P}(i, \mathcal{A}), \quad \text{where} \quad Q_f = \{\mu, j, e\} \ \text{and} \ Q_k = \{\mu, j\}.
\end{equation}

\begin{itemize}
    \item Frame Localization Query ($Q_f$): Based on the action specification $\mathcal{A}$, the parser selects key action nodes (e.g., maximum knee flexion). The parser chooses from a predefined library of kinematic measurements $\mu$ (angles, distances, or relative positions), specifying the relevant joints $j$ and the extremum criteria $e$ to locate a key frame in the motion sequence.
    \item Kinematic Measurement Query ($Q_k$): Based on the instruction $i$, the parser defines the specific physical values to be measured at the frames identified by $Q_f$. It specifies the relevant joints $j$ and the measurement type $\mu$.
\end{itemize}

The parsing process $\mathcal{P}$ operates purely on text, ensuring the queries remain independent of the motion and visual content. These queries are then executed on the reconstructed sequences $m_q$ and $m_r$ to extract kinematic measurements $\mathcal{M}$, which is represented as structured textual data, encoding precise motion characteristics in an interpretable form. Further details are provided in the \autoref{sec:appendix_parser}.

\paragraph{Kinematic Verification.}
The extracted kinematic measurements $\mathcal{M}$ are provided to the evaluated model alongside the original task inputs $(v_q, v_r, i)$. This augmentation allows the model to perform grounded reasoning through a verification mapping $\mathcal{V}$ on both visual and kinematic measurements:
\begin{equation}
y = \mathcal{V}(v_q, v_r, i, \mathcal{M}).
\end{equation}
By incorporating explicit, verifiable motion measurements $\mathcal{M}$, the model can shift from relying on ambiguous visual observations to evidence-based judgment. This integration effectively bridges the gap between visual perception and physical reality, suggesting that precise kinematic measurements can help mitigate motion hallucinations under paired-video evaluation.

\begin{table}[t]
\centering
\caption{Results of various LMMs on MotionHalluc. Models with PPV are shown in gray. Subscripts denote performance change relative to the base model (\textcolor{teal}{blue}: gain, \textcolor{red}{red}: loss). Directional Hallucination (DH) is split into Original and Reversed sequences to assess directional bias. Values are accuracy (\%). In each task, best in \textbf{bold}, second best \underline{underlined} (excluding sub-columns).}
\label{tab:main-results}

\newcommand{\gain}[1]{\textsubscript{\textcolor{teal}{\scriptsize($\uparrow$#1)}}}
\newcommand{\loss}[1]{\textsubscript{\textcolor{red}{\scriptsize($\downarrow$#1)}}}

\resizebox{\linewidth}{!}{%
\begin{tabular}{lcccccc}
\toprule
\multirow{2}{*}{\textbf{Model}} & \multicolumn{3}{c}{\textbf{DH}} & \multirow{2}{*}{\textbf{AH}} & \multirow{2}{*}{\textbf{TH}} & \multirow{2}{*}{\textbf{Avg}} \\ \cmidrule(lr){2-4}
 & Original & Reversed & Avg & & & \\ \midrule
InternVL3.5-8B & 73.07 & 41.34 & 57.21 & 57.33 & 50.31 & 54.95 \\
\rowcolor{gray!20} +PPV & 71.79 & 45.19 & 58.49 \gain{1.28} & 62.50 \gain{5.17} & 63.29 \gain{12.98} & 61.42 \gain{6.47} \\ \addlinespace[0.5em]
LLaVA-OV-1.5-8B & 98.39 & 1.92 & 50.16 & 53.50 & 41.45 & 48.37 \\
\rowcolor{gray!20} +PPV & 99.36 & 0 & 49.67 \loss{0.49} & 57.00 \gain{3.50} & 74.36 \gain{32.91} & 60.34 \gain{11.97} \\ \addlinespace[0.5em]
Qwen3-VL-8B & 61.21 & 48.39 & 54.80 & 66.16 & 62.97 & 61.31 \\
\rowcolor{gray!20} +PPV & 71.79 & 45.51 & 58.65 \gain{3.85} & 64.00 \loss{2.16} & 76.58 \gain{13.61} & 66.41 \gain{5.10} \\  \midrule \addlinespace[0.5em]
Gemini-3-Flash & 91.45 & 38.67 & 65.06 & 83.50 & 53.05 & 67.21 \\
\rowcolor{gray!20} +PPV & 81.83 & 83.54 & \underline{82.69} \gain{17.63} & \underline{89.77} \gain{6.27} & \textbf{80.37} \gain{27.32} & \textbf{84.28} \gain{17.07} \\ \addlinespace[0.5em]
Qwen3.5-plus & 90.06 & 38.14 & 64.10 & 81.72 & 68.03 & 71.28 \\ 
\rowcolor{gray!20} +PPV & 86.96 & 79.70 & \textbf{83.33} \gain{19.23} & \textbf{89.94} \gain{8.22} & \underline{77.74} \gain{9.71} & \underline{83.67} \gain{12.39} \\ 
\bottomrule
\end{tabular}%
}
\end{table}

\section{Experiments}

We conduct extensive evaluations of five representative Large Multimodal Models (LMMs) on our proposed MotionHalluc benchmark, including Gemini-3-Flash \cite{team2023gemini}, Qwen3.5-plus \cite{bai2023qwen}, and three leading open-source models,  InternVL3.5-8B \cite{wang2025internvl3_5}, LLaVA-OV-1.5-8B \cite{an2025llava} and Qwen3-VL-8B \cite{bai2023qwen}. Specially, Gemini-3-Flash is used as the semantic parser in PPV. For all evaluations, we maintain the original configurations of each model. Open-source models are evaluated with sampling disabled, using greedy decoding for deterministic outputs. In contrast, for proprietary models, we retain default generation settings and repeat each experiment three times to mitigate potential randomness, reporting the average results. Notably, our experiments follow the video processing protocol in VidDiff \cite{burgess2025video} to unify the sampling rate at FPS = 4, which effectively captures essential kinematic dynamics while maintaining computational efficiency. All experiments for open-source models were conducted on ten NVIDIA RTX3090 GPUs.

\subsection{Evaluation on MotionHalluc}

\autoref{tab:main-results} reports the comprehensive performance of all evaluated LMMs on the MotionHalluc benchmark. A critical observation emerges from the DH task, where a stark performance disparity exists between Original and Reversed video sequences. Models demonstrate high accuracy in cases where the video order follows standard corrective logic. However, a significant collapse occurs when the video order is reversed (exchange the query and reference videos), with accuracy metrics frequently falling to less than half of their original values. This discrepancy suggests that models do not perform genuine bidirectional kinematic comparison; instead, they rely heavily on the standalone semantic plausibility of the instruction or an inherent bias that the second video inherently represents a "better" execution. When this comparative context is disrupted in the reversed setting, the models lack the robustness to verify the instruction against the actual motion transformation, leading to a surge in directional hallucinations. To assess the stability of these results, we use 10,000 action-stratified cluster bootstrap resamples of the same underlying video pairs and conduct 32 leave-one-action-out evaluations (Appendix~\ref{sec:statistical_reliability}).

Regarding the AH task, the primary challenge lies in correctly identifying both the specific joints responsible for the motion discrepancy and their corresponding directions of change. Our results show that most models struggle with this task, especially when provided with semantically reasonable instructions. When faced with multiple plausible-sounding options, models frequently fail to select the one that accurately reflects the kinematic grounding within the video pairs. This fundamental deficiency in discriminative reasoning indicates that current LMMs lack the capacity for precise motion understanding and comparison. For the TH task, due to the high visual similarity within the feature space of the visual encoder, models often struggle to distinguish between the specific target action phase and other temporally adjacent segments that appear visually analogous, highlighting a fundamental weakness in fine-grained temporal reasoning for paired-video analysis.

\subsection{Evaluation of PPV on MotionHalluc}
To examine the effectiveness of PPV on the MotionHalluc benchmark, \autoref{tab:main-results} presents the results of the PPV across all evaluated LMMs, with subscript annotations indicating the gain or loss compared to the original model performance. The integration of kinematic evidence through PPV leads to consistent improvements across all three hallucination dimensions for most models. Notably, leveraging their robust reasoning capabilities, both Gemini-3-Flash and Qwen3.5-plus exhibit substantial improvements across a diverse range of tasks. The most striking improvement is observed in Temporal Hallucination, where LLaVA-OV-1.5-8B achieves a remarkable 32.91 percentage point increase, demonstrating that explicit kinematic evidence can effectively resolve temporal ambiguities. For all five models, the paired 95\% bootstrap confidence intervals for PPV-GT overall gains exclude zero, and the gains remain positive in every leave-one-action-out split within the 32 evaluated Fit3D actions. To further contextualize these gains, we examine changes in Original/Reversed DH accuracy and affirmative-response bias in Appendix~\ref{sec:dh_response_bias}.

These results underscore the critical role of kinematic grounding in enhancing the reliability of motion instruction generation and highlight PPV as a promising direction for mitigating hallucinations in LMMs. However, it is important to note that while PPV significantly reduces hallucinations, it does not completely eliminate them, indicating that further research is needed to fully address the underlying challenges in motion understanding and reasoning. PPV also incurs additional latency, primarily due to model initialization and, for uncached video pairs, motion reconstruction.  In our timing setup with 4D-Humans on a single RTX 3090, the initial startup took approximately 87~s.  After initialization, reconstructing motion for a query video and its reference video was estimated to take approximately 163~s.

\definecolor{myred}{RGB}{255,0,0}
\definecolor{myorange}{RGB}{237,125,49}
\definecolor{myyellow}{RGB}{255,255,0}
\begin{figure}[t] 
    \centering
    \includegraphics[width=\linewidth, trim=22mm 17mm 17mm 0mm, clip]{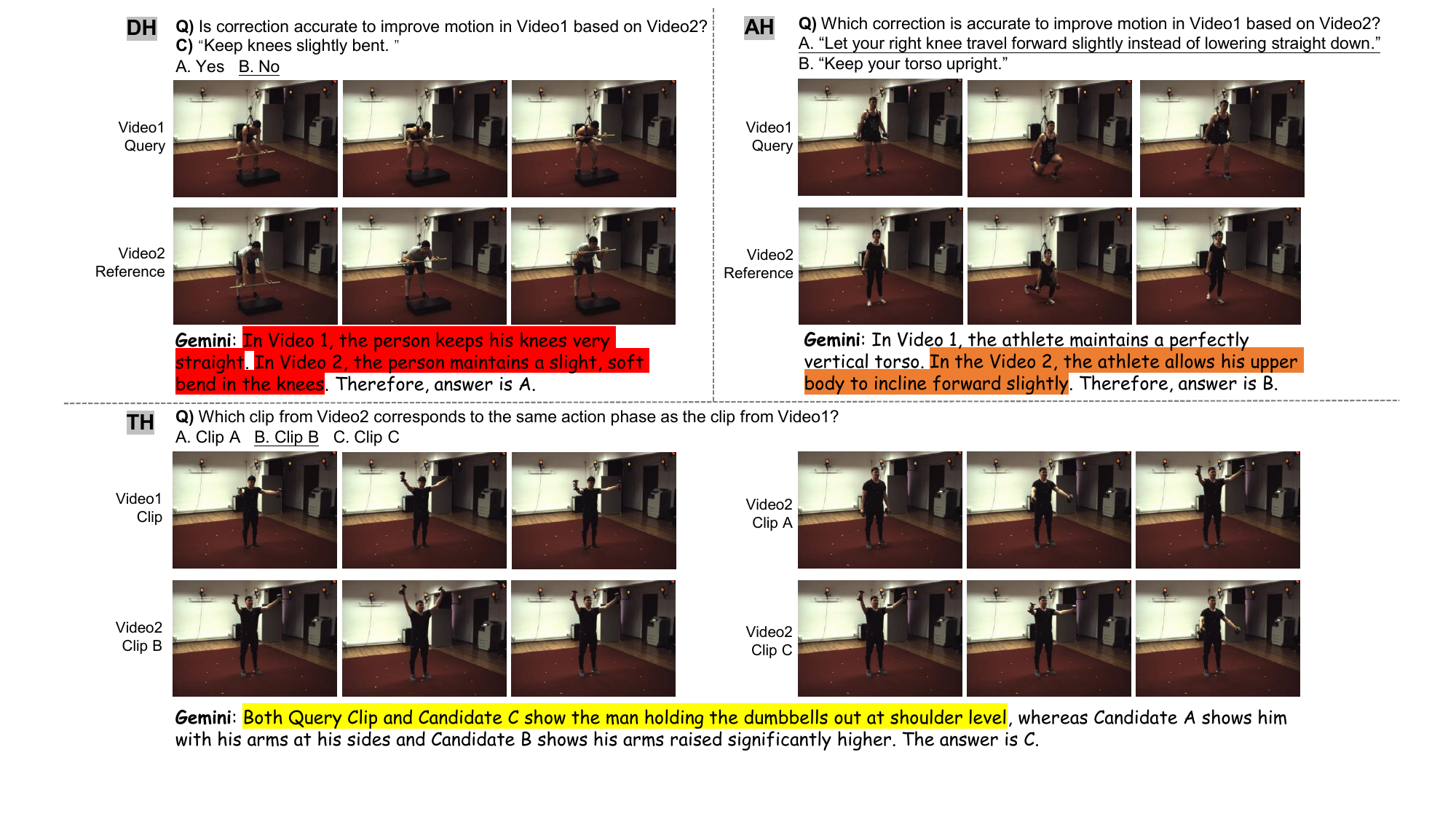} 
    \setlength{\fboxsep}{0.5pt}
    \vspace{-3mm}
    \caption{
    Failure cases. We highlight
    \fcolorbox{white}{myred}{\textbf{reversed correspondence}},
    \fcolorbox{white}{myorange}{\textbf{misattributed body part}}, and
    \fcolorbox{white}{myyellow}{\textbf{temporal alignment based on appearance similarity}}
    failure.
    Correct answers are \underline{underlined}.
    }
    \label{fig:failure_cases} 
\end{figure}

\subsection{Failure Case Analysis}

To further investigate the hallucination patterns, we re-evaluate a set of failure cases on Gemini-3-Flash \cite{team2023gemini} in a standard setting (without PPV) and ask the model to provide a brief analysis together with its final answer. As shown in \autoref{fig:failure_cases}, we observe distinct failure modes. In the DH task, the model can recognize the correct motion attributes but reverses their correspondence between the query and reference videos. In the AH task, the model assigns the instruction to an incorrect body part or pose attribute, over-committing to a visually plausible explanation even when the evidence does not support it. As for the TH task, the model describes the clips correctly at a local level but matches them to the wrong motion phase, relying on appearance similarity rather than temporal alignment. Together, these cases show that the model tends to ground its decisions in superficial cues rather than in the actual kinematic relationship between the two videos.

Overall, these failure cases confirm that current models remain vulnerable to multiple forms of motion hallucination, even when they can generate fluent and seemingly reasonable analyses. The core issue is not merely inaccurate language generation, but a deeper inability to verify whether a candidate instruction is consistent with the true motion difference.

\begin{table}[t]
\centering
\caption{Ablation study of PPV components on Qwen3.5-plus. Recon. denotes reconstructed motion, and GT denotes ground-truth motion capture data.}
\label{tab:ablation-results}

\begin{tabular}{lcccccc}
\toprule
\multirow{2}{*}{\textbf{Model}} & \multicolumn{3}{c}{\textbf{DH}} & \multirow{2}{*}{\textbf{AH}} & \multirow{2}{*}{\textbf{TH}} & \multirow{2}{*}{\textbf{Avg}} \\ \cmidrule(lr){2-4}
 & Original & Reversed & Avg & & & \\ \midrule

Qwen3.5-plus & 90.06 & 38.14 & 64.1 & 81.72 & 68.03 & 71.28 \\
+Semantic Hint & 92.84 & 35.47 & 64.21 & 80.38 & 67.72 & 70.77 \\
+PPV (Recon.) & 86.96 & 79.7 & 83.33 & 89.94 & 77.74 & 83.67 \\
+PPV (GT) & 87.71 & 81.83 & 84.77 & 90.27 & 74.05 & 83.03 \\

\bottomrule
\end{tabular}%
\end{table}

\subsection{Ablation Study on PPV}
\label{sec:ablation}
We conduct ablation experiments on Qwen3.5-plus to examine whether the performance gains arise from additional language prompting or from the introduction of kinematic measurements. We first study the necessity of kinematic measurements, and then analyze the impact of motion reconstruction accuracy. Results are shown in \autoref{tab:ablation-results}.

\paragraph{Necessity of Kinematic Measurements.}
A critical question is whether the performance gains of PPV stem from the kinematic measurements or simply from the semantic hints provided by the queries. Instead of executing queries to obtain kinematic measurements, we use a template to transform queries into textual prompts that remind the model of relevant body-part attributes (e.g., joint angles, relative positions). Our results reveal that without the numerical measurements, the model's performance degrades significantly, falling back to levels comparable to the base model. This confirms that semantic guidance alone is insufficient for resolving motion hallucinations, while the precise, quantitative kinematic measurements are the primary driver of the observed gains in the PPV-augmented evaluation setting.

\paragraph{Robustness to Motion Accuracy.}
We further evaluate PPV using Ground-Truth motion data instead of reconstructed sequences. The ablation results indicate that utilizing GT motion yields a maximum performance difference of only 3.7\% across all evaluation tasks. Notably, while GT yields slight gains in most tasks, reconstructed motion counter-intuitively observes a higher accuracy than GT in the TH task. This marginal gap suggests that PPV is not sensitive to motion accuracy. Because evidence is provided in query–reference pairs, the model focuses on relative discrepancies rather than absolute values, effectively mitigating the impact of systematic errors in motion reconstruction.

\section{Conclusion}

In this work, we identify motion hallucinations in the motion instruction task, introducing MotionHalluc as a diagnostic benchmark to quantify the physical reliability of LMMs across directional, attributional, and temporal dimensions. Our results suggest that current models' susceptibility to hallucinations is mainly associated with a reliance on linguistic priors rather than kinematic differences. To bridge this gap, we provide PPV, a training-free evidence extraction and verification baseline that augments LMMs with kinematic evidence, supporting evidence-based judgment in our evaluation setting. Our results demonstrate that explicit kinematic grounding is essential for mitigating hallucinations and achieving reliable motion analysis. 

\paragraph{Limitations and Future Work.}
Our current study primarily focuses on indoor fitness-related movements, which are typically structured and captured under controlled conditions. As a result, our benchmark does not yet cover more diverse and unconstrained outdoor activities, such as soccer or tennis, where motion patterns are more complex, involve multiple participants, and are subject to greater environmental variability. This limitation may affect the generalizability of our findings to real-world settings, where motion understanding often requires reasoning over noisier observations and more dynamic interactions. We also acknowledge potential risks of misuse, such as applying motion understanding systems in surveillance, which are beyond the intended scope of this work. Future work will extend to more diverse and unconstrained motion scenarios to better reflect real-world conditions.

\clearpage
\bibliographystyle{unsrtnat}
\bibliography{refs}

@article{hurst2024gpt,
  title={Gpt-4o system card},
  author={Hurst, Aaron and Lerer, Adam and Goucher, Adam P and Perelman, Adam and Ramesh, Aditya and Clark, Aidan and Ostrow, AJ and Welihinda, Akila and Hayes, Alan and Radford, Alec and others},
  journal={arXiv preprint arXiv:2410.21276},
  year={2024}
}

@article{an2025llava,
  title={Llava-onevision-1.5: Fully open framework for democratized multimodal training},
  author={An, Xiang and Xie, Yin and Yang, Kaicheng and Zhang, Wenkang and Zhao, Xiuwei and Cheng, Zheng and Wang, Yirui and Xu, Songcen and Chen, Changrui and Zhu, Didi and others},
  journal={arXiv preprint arXiv:2509.23661},
  year={2025}
}

@article{wang2025internvl3_5,
  title={InternVL3.5: Advancing Open-Source Multimodal Models in Versatility, Reasoning, and Efficiency},
  author={Wang, Weiyun and Gao, Zhangwei and Gu, Lixin and Pu, Hengjun and Cui, Long and Wei, Xingguang and Liu, Zhaoyang and Jing, Linglin and Ye, Shenglong and Shao, Jie and others},
  journal={arXiv preprint arXiv:2508.18265},
  year={2025}
}

@article{Qwen2.5-VL,
  title={Qwen2.5-VL Technical Report},
  author={Bai, Shuai and Chen, Keqin and Liu, Xuejing and Wang, Jialin and Ge, Wenbin and Song, Sibo and Dang, Kai and Wang, Peng and Wang, Shijie and Tang, Jun and Zhong, Humen and Zhu, Yuanzhi and Yang, Mingkun and Li, Zhaohai and Wan, Jianqiang and Wang, Pengfei and Ding, Wei and Fu, Zheren and Xu, Yiheng and Ye, Jiabo and Zhang, Xi and Xie, Tianbao and Cheng, Zesen and Zhang, Hang and Yang, Zhibo and Xu, Haiyang and Lin, Junyang},
  journal={arXiv preprint arXiv:2502.13923},
  year={2025}
}

@article{fang2024cigtime,
  title={Cigtime: Corrective instruction generation through inverse motion editing},
  author={Fang, Qihang and Tang, Chengcheng and Tekin, Bugra and Yang, Yanchao},
  journal={Advances in Neural Information Processing Systems},
  volume={37},
  pages={102011--102035},
  year={2024}
}

@inproceedings{yeh2025coachme,
  title={CoachMe: Decoding Sport Elements with a Reference-Based Coaching Instruction Generation Model},
  author={Yeh, Wei-Hsin and Su, Yu-An and Chen, Chih-Ning and Lin, Yi-Hsueh and Ku, Calvin and Chiu, Wenhsin and Hu, Min-Chun and Ku, Lun-Wei},
  booktitle={Proceedings of the 63rd Annual Meeting of the Association for Computational Linguistics (Volume 1: Long Papers)},
  pages={29126--29151},
  year={2025}
}

@article{burgess2025video,
  title={Video action differencing},
  author={Burgess, James and Wang, Xiaohan and Zhang, Yuhui and Rau, Anita and Lozano, Alejandro and Dunlap, Lisa and Darrell, Trevor and Yeung-Levy, Serena},
  journal={arXiv preprint arXiv:2503.07860},
  year={2025}
}

@article{comanici2025gemini,
  title={Gemini 2.5: Pushing the frontier with advanced reasoning, multimodality, long context, and next generation agentic capabilities},
  author={Comanici, Gheorghe and Bieber, Eric and Schaekermann, Mike and Pasupat, Ice and Sachdeva, Noveen and Dhillon, Inderjit and Blistein, Marcel and Ram, Ori and Zhang, Dan and Rosen, Evan and others},
  journal={arXiv preprint arXiv:2507.06261},
  year={2025}
}

@inproceedings{li2025unipose,
  title={Unipose: A unified multimodal framework for human pose comprehension, generation and editing},
  author={Li, Yiheng and Hou, Ruibing and Chang, Hong and Shan, Shiguang and Chen, Xilin},
  booktitle={Proceedings of the Computer Vision and Pattern Recognition Conference},
  pages={27805--27815},
  year={2025}
}

@inproceedings{delmas2023posefix,
  title={Posefix: Correcting 3d human poses with natural language},
  author={Delmas, Ginger and Weinzaepfel, Philippe and Moreno-Noguer, Francesc and Rogez, Gr{\'e}gory},
  booktitle={Proceedings of the IEEE/CVF International Conference on Computer Vision},
  pages={15018--15028},
  year={2023}
}

@inproceedings{verma2020yoga,
  title={Yoga-82: a new dataset for fine-grained classification of human poses},
  author={Verma, Manisha and Kumawat, Sudhakar and Nakashima, Yuta and Raman, Shanmuganathan},
  booktitle={Proceedings of the IEEE/CVF conference on computer vision and pattern recognition workshops},
  pages={1038--1039},
  year={2020}
}

@article{anand2022yoga,
  title={Yoga pose estimation and feedback generation using deep learning},
  author={Anand Thoutam, Vivek and Srivastava, Anugrah and Badal, Tapas and Kumar Mishra, Vipul and Sinha, GR and Sakalle, Aditi and Bhardwaj, Harshit and Raj, Manish},
  journal={Computational Intelligence and Neuroscience},
  volume={2022},
  number={1},
  pages={4311350},
  year={2022},
  publisher={Wiley Online Library}
}

@inproceedings{parmar2022domain,
  title={Domain knowledge-informed self-supervised representations for workout form assessment},
  author={Parmar, Paritosh and Gharat, Amol and Rhodin, Helge},
  booktitle={European conference on computer vision},
  pages={105--123},
  year={2022},
  organization={Springer}
}

@article{wang2026integrating,
  title={Integrating multimodal AI technologies for sports injury prediction and rehabilitation: Systematic review},
  author={Wang, Pengbo and Wang, Aodi and Wang, Saidi},
  journal={Journal of Human Sport and Exercise},
  volume={21},
  number={1},
  pages={22--37},
  year={2026}
}

@article{du2025motionsight,
  title={Motionsight: Boosting fine-grained motion understanding in multimodal llms},
  author={Du, Yipeng and Fan, Tiehan and Nan, Kepan and Xie, Rui and Zhou, Penghao and Li, Xiang and Yang, Jian and Yang, Zhenheng and Tai, Ying},
  journal={arXiv preprint arXiv:2506.01674},
  year={2025}
}

@inproceedings{ye2026mm,
  title={Mm-spubench: Towards better understanding of spurious biases in multimodal llms},
  author={Ye, Wenqian and Liu, Bohan and Zheng, Guangtao and Wang, Di and Cao, Xu and Ma, Yunsheng and Lai, Bolin and Rehg, James M and Zhang, Aidong},
  booktitle={Proceedings of the 32nd ACM SIGKDD Conference on Knowledge Discovery and Data Mining V. 1},
  pages={2854--2865},
  year={2026}
}

@inproceedings{han2024instinctive,
  title={The instinctive bias: Spurious images lead to illusion in mllms},
  author={Han, Tianyang and Lian, Qing and Pan, Rui and Pi, Renjie and Zhang, Jipeng and Diao, Shizhe and Lin, Yong and Zhang, Tong},
  booktitle={Proceedings of the 2024 Conference on Empirical Methods in Natural Language Processing},
  pages={16163--16177},
  year={2024}
}

@article{nie2024slowfocus,
  title={Slowfocus: Enhancing fine-grained temporal understanding in video llm},
  author={Nie, Ming and Ding, Dan and Wang, Chunwei and Guo, Yuanfan and Han, Jianhua and Xu, Hang and Zhang, Li},
  journal={Advances in Neural Information Processing Systems},
  volume={37},
  pages={81808--81835},
  year={2024}
}

@inproceedings{li2025vidhalluc,
  title={Vidhalluc: Evaluating temporal hallucinations in multimodal large language models for video understanding},
  author={Li, Chaoyu and Im, Eun Woo and Fazli, Pooyan},
  booktitle={Proceedings of the IEEE/CVF Conference on Computer Vision and Pattern Recognition},
  pages={13723--13733},
  year={2025}
}

@inproceedings{kong2025mhbench,
  title={Mhbench: Demystifying motion hallucination in videollms},
  author={Kong, Ming and Zeng, Xianzhou and Chen, Luyuan and Li, Yadong and Yan, Bo and Zhu, Qiang},
  booktitle={Proceedings of the AAAI Conference on Artificial Intelligence},
  volume={39},
  pages={4401--4409},
  year={2025}
}

@article{li2025videohallu,
  title={Videohallu: Evaluating and mitigating multi-modal hallucinations on synthetic video understanding},
  author={Li, Zongxia and Wu, Xiyang and Shi, Guangyao and Qin, Yubin and Du, Hongyang and Liu, Fuxiao and Zhou, Tianyi and Manocha, Dinesh and Boyd-Graber, Jordan Lee},
  journal={arXiv preprint arXiv:2505.01481},
  year={2025}
}

@inproceedings{parmar2019and,
  title={What and how well you performed? a multitask learning approach to action quality assessment},
  author={Parmar, Paritosh and Morris, Brendan Tran},
  booktitle={Proceedings of the IEEE/CVF conference on computer vision and pattern recognition},
  pages={304--313},
  year={2019}
}

@inproceedings{pirsiavash2014assessing,
  title={Assessing the quality of actions},
  author={Pirsiavash, Hamed and Vondrick, Carl and Torralba, Antonio},
  booktitle={European conference on computer vision},
  pages={556--571},
  year={2014},
  organization={Springer}
}

@inproceedings{wu2021computer,
  title={A computer vision-based yoga pose grading approach using contrastive skeleton feature representations},
  author={Wu, Yubin and Lin, Qianqian and Yang, Mingrun and Liu, Jing and Tian, Jing and Kapil, Dev and Vanderbloemen, Laura},
  booktitle={Healthcare},
  volume={10},
  pages={36},
  year={2021},
  organization={MDPI}
}

@inproceedings{li20223d,
  title={3D-Yoga: a 3D yoga dataset for visual-based hierarchical sports action analysis},
  author={Li, Jianwei and Hu, Haiqing and Li, Jinyang and Zhao, Xiaomei},
  booktitle={Proceedings of the Asian Conference on Computer Vision},
  pages={434--450},
  year={2022}
}

@article{dong2024interpretable,
  title={Interpretable long-term action quality assessment},
  author={Dong, Xu and Liu, Xinran and Li, Wanqing and Adeyemi-Ejeye, Anthony and Gilbert, Andrew},
  journal={arXiv preprint arXiv:2408.11687},
  year={2024}
}

@article{henriques2026can,
  title={Can Vision Language Models Judge Action Quality? An Empirical Evaluation},
  author={Henriques, Rui and Rei, Ricardo and Martins, Pedro Henrique and others},
  journal={arXiv preprint arXiv:2604.08294},
  year={2026}
}

@article{wang2025attention,
  title={Attention-driven multimodal alignment for long-term action quality assessment},
  author={Wang, Xin and Li, Peng-Jie and Shen, Yuan-Yuan},
  journal={Applied Soft Computing},
  pages={113649},
  year={2025},
  publisher={Elsevier}
}

@article{tharatipyakul2024deep,
  title={Deep learning-based human body pose estimation in providing feedback for physical movement: A review},
  author={Tharatipyakul, Atima and Srikaewsiew, Thanawat and Pongnumkul, Suporn},
  journal={Heliyon},
  volume={10},
  number={17},
  year={2024},
  publisher={Elsevier}
}

@inproceedings{li2023evaluating,
  title={Evaluating object hallucination in large vision-language models},
  author={Li, Yifan and Du, Yifan and Zhou, Kun and Wang, Jinpeng and Zhao, Xin and Wen, Ji-Rong},
  booktitle={Proceedings of the 2023 conference on empirical methods in natural language processing},
  pages={292--305},
  year={2023}
}

@inproceedings{papineni2002bleu,
  title={Bleu: a method for automatic evaluation of machine translation},
  author={Papineni, Kishore and Roukos, Salim and Ward, Todd and Zhu, Wei-Jing},
  booktitle={Proceedings of the 40th annual meeting of the Association for Computational Linguistics},
  pages={311--318},
  year={2002}
}

@inproceedings{lin2004rouge,
  title={Rouge: A package for automatic evaluation of summaries},
  author={Lin, Chin-Yew},
  booktitle={Text summarization branches out},
  pages={74--81},
  year={2004}
}

@article{liu2023g,
  title={G-eval: Nlg evaluation using gpt-4 with better human alignment, 2023},
  author={Liu, Yang and Iter, Dan and Xu, Yichong and Wang, Shuohang and Xu, Ruochen and Zhu, Chenguang},
  journal={arXiv preprint arXiv:2303.16634},
  volume={12},
  pages={1},
  year={2023}
}

@article{gao2025exploring,
  title={Exploring hallucination of large multimodal models in video understanding: Benchmark, analysis and mitigation},
  author={Gao, Hongcheng and Qu, Jiashu and Tang, Jingyi and Bi, Baolong and Liu, Yue and Chen, Hongyu and Liang, Li and Su, Li and Huang, Qingming},
  journal={arXiv preprint arXiv:2503.19622},
  year={2025}
}

@article{liberatori2025convis,
  title={ConViS-Bench: Estimating Video Similarity Through Semantic Concepts},
  author={Liberatori, Benedetta and Conti, Alessandro and Vaquero, Lorenzo and Wang, Yiming and Ricci, Elisa and Rota, Paolo},
  journal={arXiv preprint arXiv:2509.19245},
  year={2025}
}

@article{DBLP:journals/tog/LoperM0PB15,
  author       = {Matthew Loper and
                  Naureen Mahmood and
                  Javier Romero and
                  Gerard Pons{-}Moll and
                  Michael J. Black},
  title        = {{SMPL:} a skinned multi-person linear model},
  journal      = {{ACM} Trans. Graph.},
  volume       = {34},
  number       = {6},
  pages        = {248:1--248:16},
  year         = {2015},
  url          = {https://doi.org/10.1145/2816795.2818013},
  doi          = {10.1145/2816795.2818013},
  bibsource    = {dblp computer science bibliography, https://dblp.org}
}

@inproceedings{chen2025skeleton,
  title={Skeleton-based action recognition with non-linear dependency modeling and hilbert-schmidt independence criterion},
  author={Chen, Haipeng and Yang, Yuheng and Lyu, Yingda},
  booktitle={Proceedings of the AAAI Conference on Artificial Intelligence},
  volume={39},
  pages={2043--2051},
  year={2025}
}

@article{liu2022end,
  title={End-to-end temporal action detection with transformer},
  author={Liu, Xiaolong and Wang, Qimeng and Hu, Yao and Tang, Xu and Zhang, Shiwei and Bai, Song and Bai, Xiang},
  journal={IEEE Transactions on Image Processing},
  volume={31},
  pages={5427--5441},
  year={2022},
  publisher={IEEE}
}

@inproceedings{goel2023humans,
  title={Humans in 4d: Reconstructing and tracking humans with transformers},
  author={Goel, Shubham and Pavlakos, Georgios and Rajasegaran, Jathushan and Kanazawa, Angjoo and Malik, Jitendra},
  booktitle={Proceedings of the IEEE/CVF International Conference on Computer Vision},
  pages={14783--14794},
  year={2023}
}

@inproceedings{shin2024wham,
  title={Wham: Reconstructing world-grounded humans with accurate 3d motion},
  author={Shin, Soyong and Kim, Juyong and Halilaj, Eni and Black, Michael J},
  booktitle={Proceedings of the IEEE/CVF Conference on Computer Vision and Pattern Recognition},
  pages={2070--2080},
  year={2024}
}

@article{jiang2023motiongpt,
  title={Motiongpt: Human motion as a foreign language},
  author={Jiang, Biao and Chen, Xin and Liu, Wen and Yu, Jingyi and Yu, Gang and Chen, Tao},
  journal={Advances in Neural Information Processing Systems},
  volume={36},
  pages={20067--20079},
  year={2023}
}

@article{chen2025motionllm,
  title={Motionllm: Understanding human behaviors from human motions and videos},
  author={Chen, Ling-Hao and Lu, Shunlin and Zeng, Ailing and Zhang, Hao and Wang, Benyou and Zhang, Ruimao and Zhang, Lei},
  journal={IEEE Transactions on Pattern Analysis and Machine Intelligence},
  year={2025},
  publisher={IEEE}
}

@inproceedings{zhang2024motiongpt,
  title={Motiongpt: Finetuned llms are general-purpose motion generators},
  author={Zhang, Yaqi and Huang, Di and Liu, Bin and Tang, Shixiang and Lu, Yan and Chen, Lu and Bai, Lei and Chu, Qi and Yu, Nenghai and Ouyang, Wanli},
  booktitle={Proceedings of the AAAI Conference on Artificial Intelligence},
  volume={38},
  pages={7368--7376},
  year={2024}
}

@article{zhu2506motiongpt3,
  title={Motiongpt3: Human motion as a second modality},
  author={Zhu, Bingfan and Jiang, Biao and Wang, Sunyi and Tang, Shixiang and Chen, Tao and Luo, Linjie and Zheng, Youyi and Chen, Xin},
  journal={URL https://arxiv. org/abs/2506.24086},
  year={2025}
}

@article{hu2025hmvlm,
  title={HMVLM: Human Motion-Vision-Lanuage Model via MoE LoRA},
  author={Hu, Lei and Ye, Yongjing and Xia, Shihong},
  journal={arXiv preprint arXiv:2511.01463},
  year={2025}
}

@article{wu2024motion,
  title={Motion-agent: A conversational framework for human motion generation with llms},
  author={Wu, Qi and Zhao, Yubo and Wang, Yifan and Liu, Xinhang and Tai, Yu-Wing and Tang, Chi-Keung},
  journal={arXiv preprint arXiv:2405.17013},
  year={2024}
}

@inproceedings{fieraru2021aifit,
  title={Aifit: Automatic 3d human-interpretable feedback models for fitness training},
  author={Fieraru, Mihai and Zanfir, Mihai and Pirlea, Silviu Cristian and Olaru, Vlad and Sminchisescu, Cristian},
  booktitle={Proceedings of the IEEE/CVF conference on computer vision and pattern recognition},
  pages={9919--9928},
  year={2021}
}

@inproceedings{pavlakos2019expressive,
  title={Expressive body capture: 3d hands, face, and body from a single image},
  author={Pavlakos, Georgios and Choutas, Vasileios and Ghorbani, Nima and Bolkart, Timo and Osman, Ahmed AA and Tzionas, Dimitrios and Black, Michael J},
  booktitle={Proceedings of the IEEE/CVF conference on computer vision and pattern recognition},
  pages={10975--10985},
  year={2019}
}

@InProceedings{Guo_2022_CVPR,
    author    = {Guo, Chuan and Zou, Shihao and Zuo, Xinxin and Wang, Sen and Ji, Wei and Li, Xingyu and Cheng, Li},
    title     = {Generating Diverse and Natural 3D Human Motions From Text},
    booktitle = {Proceedings of the IEEE/CVF Conference on Computer Vision and Pattern Recognition (CVPR)},
    month     = {June},
    year      = {2022},
    pages     = {5152-5161}
}

@article{sakoe1978dynamic,
  title={Dynamic programming algorithm optimization for spoken word recognition},
  author={Sakoe, Hiroaki and Chiba, Seibi},
  journal={IEEE transactions on acoustics, speech, and signal processing},
  volume={26},
  number={1},
  pages={43--49},
  year={1978},
  publisher={IEEE}
}

@article{singh2025openai,
  title={Openai gpt-5 system card},
  author={Singh, Aaditya and Fry, Adam and Perelman, Adam and Tart, Adam and Ganesh, Adi and El-Kishky, Ahmed and McLaughlin, Aidan and Low, Aiden and Ostrow, AJ and Ananthram, Akhila and others},
  journal={arXiv preprint arXiv:2601.03267},
  year={2025}
}

@article{team2023gemini,
  title={Gemini: a family of highly capable multimodal models},
  author={Team, Gemini and Anil, Rohan and Borgeaud, Sebastian and Alayrac, Jean-Baptiste and Yu, Jiahui and Soricut, Radu and Schalkwyk, Johan and Dai, Andrew M and Hauth, Anja and Millican, Katie and others},
  journal={arXiv preprint arXiv:2312.11805},
  year={2023}
}

@article{bai2023qwen,
  title={Qwen technical report},
  author={Bai, Jinze and Bai, Shuai and Chu, Yunfei and Cui, Zeyu and Dang, Kai and Deng, Xiaodong and Fan, Yang and Ge, Wenbin and Han, Yu and Huang, Fei and others},
  journal={arXiv preprint arXiv:2309.16609},
  year={2023}
}

@inproceedings{bae2025mash,
  title={Mash-vlm: Mitigating action-scene hallucination in video-llms through disentangled spatial-temporal representations},
  author={Bae, Kyungho and Kim, Jinhyung and Lee, Sihaeng and Lee, Soonyoung and Lee, Gunhee and Choi, Jinwoo},
  booktitle={Proceedings of the Computer Vision and Pattern Recognition Conference},
  pages={13744--13753},
  year={2025}
}

@article{luo2025dr,
  title={Dr. V: A Hierarchical Perception-Temporal-Cognition Framework to Diagnose Video Hallucination by Fine-grained Spatial-Temporal Grounding},
  author={Luo, Meng and Wu, Shengqiong and Jing, Liqiang and Ju, Tianjie and Zheng, Li and Lai, Jinxiang and Wu, Tianlong and Du, Xinya and Li, Jian and Yan, Siyuan and others},
  journal={arXiv preprint arXiv:2509.11866},
  year={2025}
}

@article{wu2025mass,
  title={MASS: Motion-Aware Spatial-Temporal Grounding for Physics Reasoning and Comprehension in Vision-Language Models},
  author={Wu, Xiyang and Li, Zongxia and Jin, Jihui and Shi, Guangyao and KV, Gouthaman and Raj, Vishnu and Sinha, Nilotpal and Chen, Jingxi and Du, Fan and Manocha, Dinesh},
  journal={arXiv preprint arXiv:2511.18373},
  year={2025}
}

@article{ionescu2013human3,
  title={Human3. 6m: Large scale datasets and predictive methods for 3d human sensing in natural environments},
  author={Ionescu, Catalin and Papava, Dragos and Olaru, Vlad and Sminchisescu, Cristian},
  journal={IEEE transactions on pattern analysis and machine intelligence},
  volume={36},
  number={7},
  pages={1325--1339},
  year={2013},
  publisher={IEEE}
}

\clearpage
\appendix

\section{Benchmark Download Instructions}
\label{sec:bench_url}

The MotionHalluc code is available in the \href{https://github.com/Leocoding0971/MotionHalluc}{GitHub repository}, and the benchmark data are available in the \href{https://huggingface.co/datasets/leosrepo/MotionHalluc}{Hugging Face dataset repository}.

The Hugging Face dataset repository provides the benchmark data used in our experiments. Specifically, it includes: (i) the full set of benchmark annotations for all tasks (AH, TH, and DH), and (ii) the original human-written annotations used to construct the benchmark.

Since we do not own the original video data, the raw videos are not redistributed. Instructions for obtaining the source videos and preparing the data are provided in the GitHub repository, together with detailed usage guidelines and scripts for data preprocessing and evaluation.

In addition to supporting the experiments presented in this work, we release these resources to facilitate future research on fine-grained motion reasoning and hallucination analysis in multimodal systems.

\section{Experimental Configurations}
\label{sec:reproducibility}

Table~\ref{tab:reproducibility_config} identifies the semantic parser and evaluated models, and lists the Hugging Face repositories for the three local Instruct models. The local models are evaluated with sampling disabled (\texttt{do\_sample=False}). The API-based evaluated models are \texttt{gemini-3-flash-preview} and \texttt{qwen3.5-plus}.

\begingroup
\setlength{\tabcolsep}{4pt}
\renewcommand{\arraystretch}{1.15}
\begin{longtable}{@{}>{\raggedright\arraybackslash}p{0.19\linewidth}>{\raggedright\arraybackslash}p{0.31\linewidth}>{\raggedright\arraybackslash}p{\dimexpr0.50\linewidth-16pt\relax}@{}}
\caption{Semantic parser and evaluated model configurations.}
\label{tab:reproducibility_config}\\
\toprule
\textbf{Component} & \textbf{Model} & \textbf{Source} \\
\midrule
\endfirsthead
\multicolumn{3}{@{}l@{}}{\tablename~\thetable{} (continued)}\\
\toprule
\textbf{Component} & \textbf{Model / configuration} & \textbf{Source / version information} \\
\midrule
\endhead
\midrule
\multicolumn{3}{r@{}}{Continued on next page}\\
\endfoot
\bottomrule
\endlastfoot
Semantic parser & \path{gemini-3-flash-preview} & \makebox[\linewidth][c]{/} \\
\addlinespace
Gemini-3-Flash & \path{gemini-3-flash-preview} & \makebox[\linewidth][c]{/} \\
\addlinespace
Qwen3.5-plus & \path{qwen3.5-plus} & \makebox[\linewidth][c]{/} \\
\addlinespace
InternVL3.5-8B & \path{OpenGVLab/InternVL3_5-8B-Instruct} & \url{https://huggingface.co/OpenGVLab/InternVL3_5-8B-Instruct} \\
\addlinespace
LLaVA-OV-1.5-8B & \path{lmms-lab/LLaVA-OneVision-1.5-8B-Instruct} & \url{https://huggingface.co/lmms-lab/LLaVA-OneVision-1.5-8B-Instruct} \\
\addlinespace
Qwen3-VL-8B & \path{Qwen/Qwen3-VL-8B-Instruct} & \url{https://huggingface.co/Qwen/Qwen3-VL-8B-Instruct} \\
\end{longtable}
\endgroup

We report these model identifiers, repository links, and decoding settings to support reproducibility of the benchmark evaluation and PPV pipeline. For API-based models, the reported identifiers alone do not guarantee a fixed backend configuration over time, particularly for the Gemini preview model. The reported results therefore correspond to the semantic parser and model configurations documented here.

\begin{figure}[h]  
  \centering         
  \vspace{5pt}
  \includegraphics[width=\linewidth]{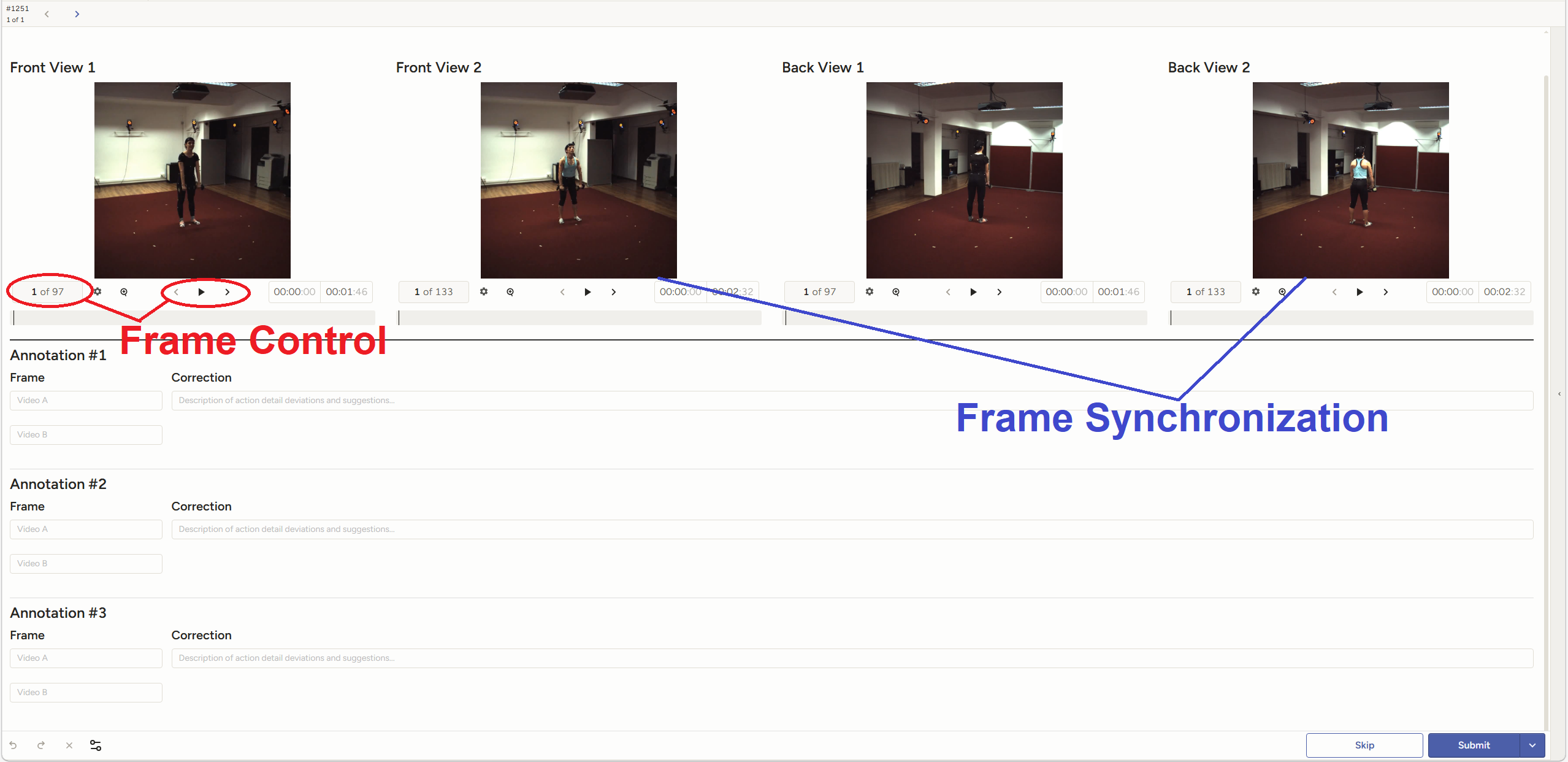}  
  \caption{Annotation interface for the MotionHalluc benchmark. The video player is specially designed to support fine-grained, frame-level comparison between query and reference motions.}  
  \label{fig:interface}           
\end{figure}

\section{Annotation Interface}

To ensure high-quality annotations for the MotionHalluc benchmark, we developed a custom annotation interface that allows annotators to efficiently and accurately identify motion differences between paired videos (\autoref{fig:interface}). For each video pair, annotators are presented with two subjects, each shown with two synchronized views (front and back). This results in four aligned video streams in total, allowing annotators to compare motions both across subjects and across viewpoints. The two views are temporally aligned at the frame level, and the interface supports frame-by-frame navigation, including precise forward and backward stepping, enabling annotators to inspect motion details with high fidelity. These design choices are intended to improve the accuracy and consistency of the annotations.

For each video pair, annotators are allowed to provide up to three corrective instructions. In practice, we observe that 48\%, 43\%, and 9\% of the video pairs contain one, two, and three corrective instructions, respectively. This distribution suggests that while the motion pairs exhibit meaningful differences, they are not excessively divergent, allowing annotators to describe discrepancies in a controlled and focused manner. Such a balance is important for constructing a benchmark that emphasizes fine-grained motion distinctions while avoiding trivial or overly ambiguous comparisons.

\begin{figure}[t] 
    \centering
    \includegraphics[width=\linewidth, trim=5mm 60mm 5mm 0mm, clip]{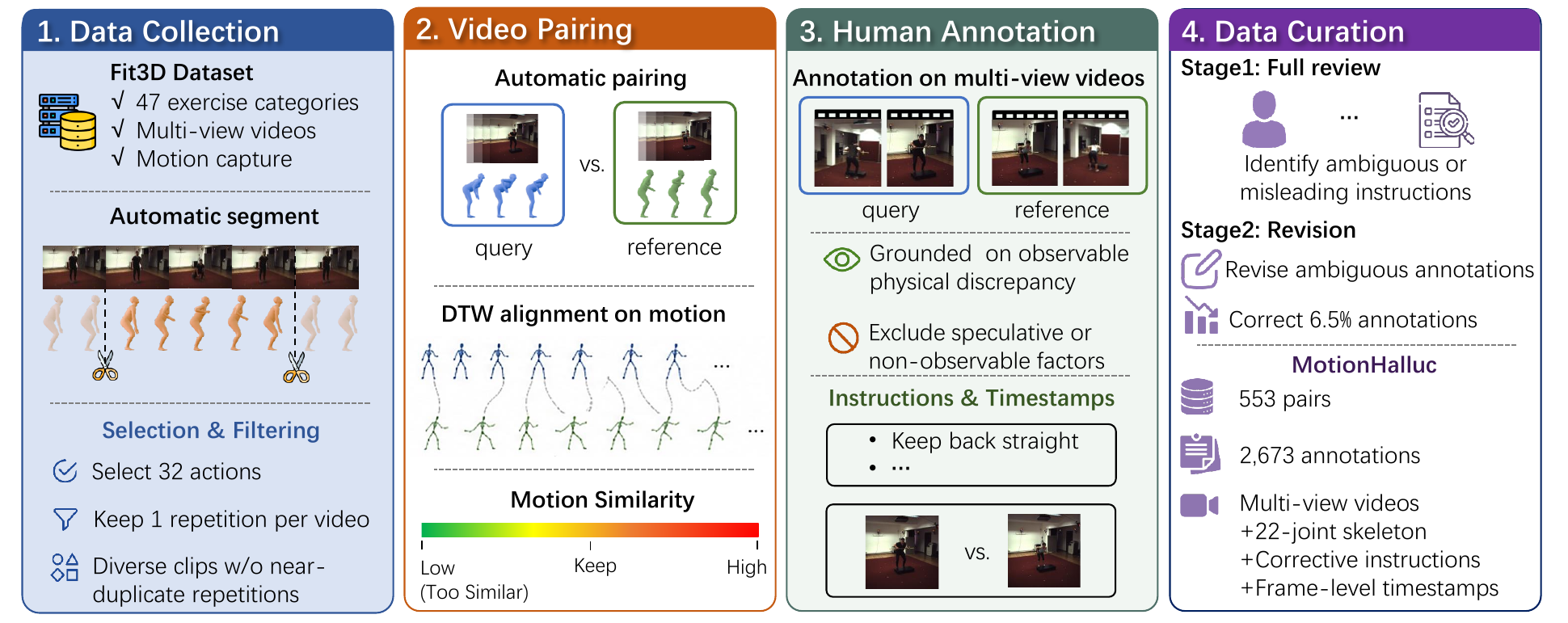} 
    \setlength{\fboxsep}{0.5pt}
    \vspace{-3mm}
    \caption{Data collection, pairing, and annotation pipeline for MotionHalluc.}
    \label{fig:data_process} 
\end{figure}

\begin{figure}[t] 
    \centering
    \includegraphics[width=\linewidth, trim=5mm 65mm 5mm 0mm, clip]{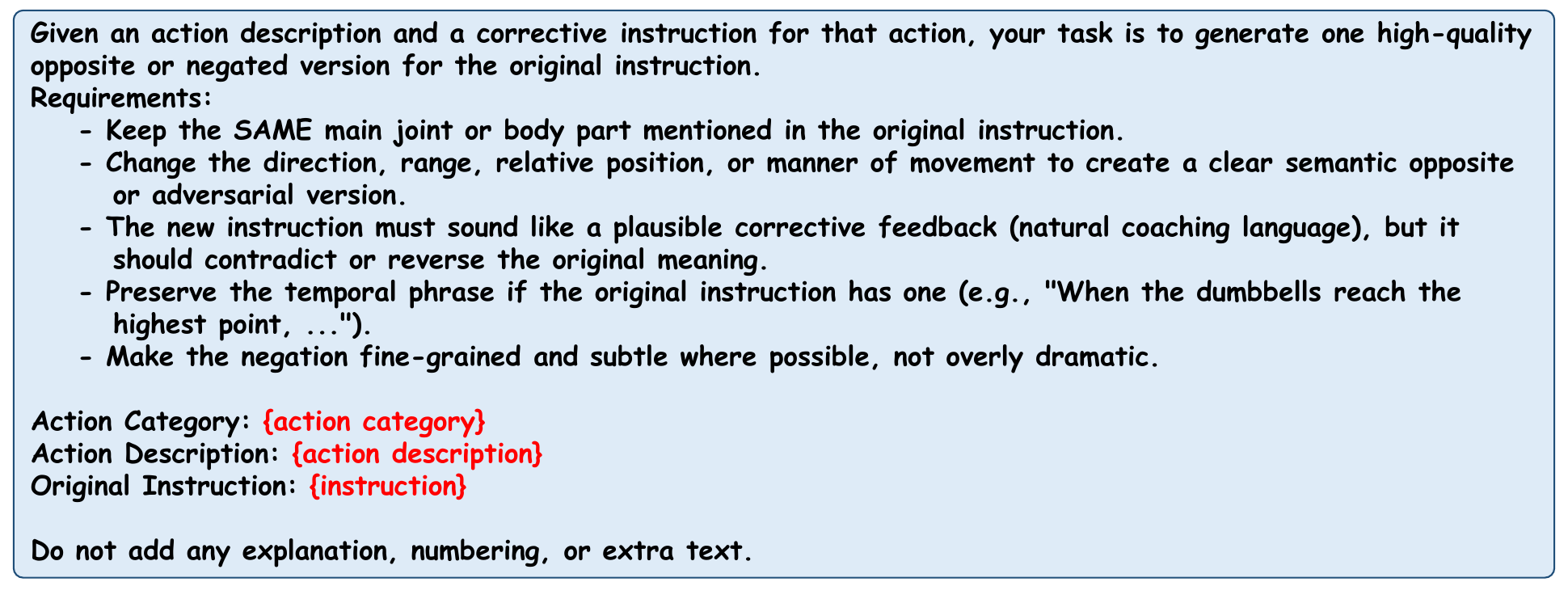} 
    \setlength{\fboxsep}{0.5pt}
    \vspace{-5mm}
    \caption{Prompt for generating semantically opposite instructions.}
    \label{fig:oppo_prompt} 
\end{figure}

\section{Human Annotation and Verification}
\label{sec:human_annotation}

The overall data collection, pairing, and annotation pipeline has been described in \autoref{sec:data_collection}, with \autoref{fig:data_process} illustrating the full workflow. Here, we provide a more detailed description of the annotation and verification procedures.

To ensure annotation quality, the primary annotation process is conducted by an annotator with prior experience in sports training. Given a pair of videos, the annotator is asked to identify visually observable motion differences, localize the corresponding frame pairs (i.e., query and reference frames), and provide corrective instructions that precisely describe the discrepancy. 

During the video pairing stage, candidate pairs are first constructed using motion similarity computed from motion representations, which serves as an initial filtering step. During annotation, the annotator is allowed to flag video pairs where no meaningful visual difference can be reliably identified. After data collection, we further perform a post-hoc filtering step by removing samples with abnormally short annotation durations, which may indicate that annotations were produced based on prior experience rather than careful inspection of the video content. Starting from 896 candidate pairs, this process removes approximately 38\% of the data, resulting in a final set of 553 video pairs with high-quality annotations.

To further improve reliability, a second annotator independently reviews the annotations, identifying ambiguous or imprecise corrective instructions. Disagreements are resolved through discussion and revision, leading to modifications in 6.5\% of the annotations. Finally, we conduct an additional quality check by randomly sampling 140 annotated instances, evenly distributed across 32 action categories, and assigning them to a third annotator for independent verification. This yields an inter-annotator disagreement rate of 2.85\%, indicating a high level of consistency across annotators.

Overall, these multi-stage annotation and verification procedures ensure that the resulting benchmark is both reliable and precise, providing a strong foundation for evaluating fine-grained motion reasoning and hallucination in multimodal systems.

\section{QA Construction}

The construction of the three QA tasks has been described in detail in \autoref{sec:evaluation_protocol}. Here, we present the prompt used to generate semantically opposite instructions (shown in \autoref{fig:oppo_prompt}). Specifically, given an action description and a corrective instruction, we prompt a language model to produce a fine-grained, plausible corrective instruction that preserves the original structure but semantically reverses the intended motion. We use GPT-5 to generate opposite instructions, and all distractors retained in the AH task have passed human verification.

We further examine the distribution of answer options across the constructed QA tasks. For DH and AH, the binary choices (A/B) are evenly distributed, with each option accounting for 50\% of the samples. For TH, which involves three-way choices, the distribution of options (A/B/C) is 31\%, 33\%, and 36\%, respectively. This near-uniform distribution indicates that the answer space is well-balanced and avoids bias toward any particular option, supporting a fair and reliable evaluation of model performance.

\begin{figure}[h] 
    \centering
    \includegraphics[width=\linewidth, trim=5mm 5mm 5mm 0mm, clip]{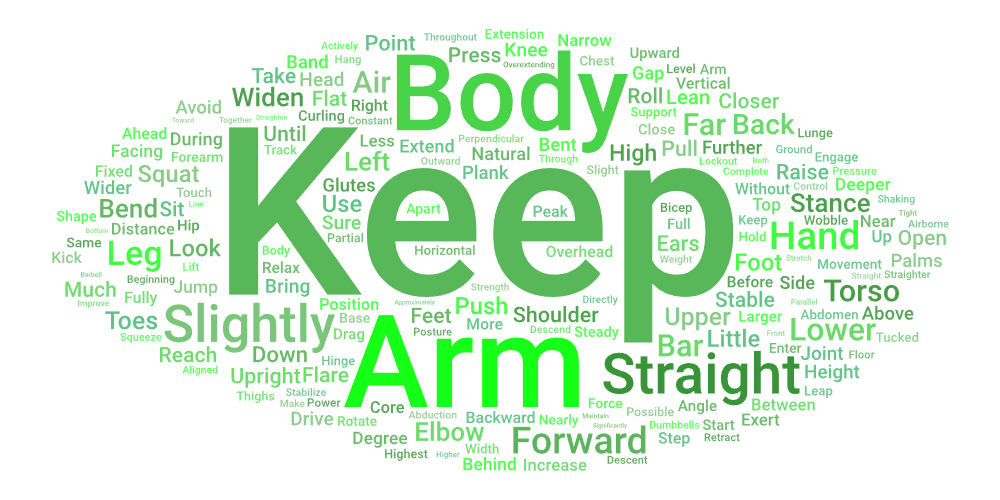} 
    \setlength{\fboxsep}{0.5pt}
    \vspace{-3mm}
    \caption{Word cloud of annotations in MotionHalluc.}
    \label{fig:word_cloud} 
\end{figure}

\section{Additional Dataset Analysis}

To provide further insights into the characteristics of our dataset, we present additional analyses of the human annotations and action categories. First, we visualize the distribution of language used in the human-written corrective instructions through a word cloud (\autoref{fig:word_cloud}). The visualization highlights frequently used terms related to body parts, spatial relations, and motion dynamics, such as joint-specific references and directional cues. This observation confirms that the annotations focus on fine-grained kinematic details and reflect consistent use of structured, instruction-like language grounded in physical motion. Second, we provide a comprehensive overview of all 32 action categories included in the dataset, along with their detailed descriptions (\autoref{tab:action_details}). These categories span a diverse set of human motions, covering various movement patterns and body configurations. The detailed descriptions further clarify the semantic scope of each category, ensuring consistent interpretation during annotation and evaluation.


\begingroup
\setlength{\tabcolsep}{3pt}
\renewcommand{\arraystretch}{1.15}
\sloppy

\begin{longtable}{|
  >{\raggedright\arraybackslash}p{0.30\linewidth}|
  >{\raggedright\arraybackslash}p{0.66\linewidth}|
}
\caption{Action categories and their detailed descriptions used in our experiments.}
\label{tab:action_details}\\
\hline
\textbf{Action category} & \textbf{Detailed description} \\
\hline
\endfirsthead

\hline
\textbf{Action category} & \textbf{Detailed description} \\
\hline
\endhead

\hline
\endfoot

\hline
\endlastfoot

band pull apart & A person stands upright, holds a resistance band at shoulder level, and pulls it outward across the chest with extended arms \\
\hline
barbell dead row & A person bends forward at the hips, holds a barbell with both hands, and pulls it upward toward the torso while keeping the back straight \\
\hline
barbell row & A person bends forward at the hips, holds a barbell with both hands, and pulls it toward the torso in a rowing motion \\
\hline
deadlift & A person lifts a barbell from the floor to a standing position by extending the hips and knees while keeping the back straight \\
\hline
diamond pushup & A person performs a push-up with hands placed close together under the chest, forming a diamond shape with the fingers \\
\hline
dumbbell biceps curls & A person holds dumbbells at the sides and curls them upward by bending the elbows while keeping the upper arms stationary \\
\hline
dumbbell hammer curls & A person holds dumbbells with a neutral grip and curls them upward by bending the elbows while keeping the upper arms stationary \\
\hline
dumbbell high pulls & A person lifts dumbbells upward toward the shoulders by pulling with the arms and raising the elbows outward \\
\hline
dumbbell overhead shoulder press & A person presses dumbbells upward from shoulder height to overhead by extending the arms while standing \\
\hline
dumbbell reverse lunge & A person steps backward into a lunge while holding dumbbells at the sides, lowering the back knee toward the floor and keeping the torso upright \\
\hline
dumbbell scaptions & A person raises dumbbells in a V-shape in front of the body with straight arms, lifting them to shoulder height \\
\hline
mule kick & A person supports the body on both hands and feet, and kicks both legs backward into the air \\
\hline
one arm row & A person stands, holds a dumbbell in the right hand, keeps the left arm straight, and pulls the dumbbell toward the torso \\
\hline
overhead trap raises & A person raises a dumbbell overhead in front of the body, lifting the shoulders and arms upward while keeping the arms straight \\
\hline
pushup & A person lowers and raises the body by bending and extending the arms while keeping the body straight in a plank position \\
\hline
barbell shrug & A person holds a barbell at thigh level and lifts the shoulders upward toward the ears, then lowers them back down \\
\hline
side lateral raise & A person raises dumbbells outward to the sides from the hips to shoulder height with straight or slightly bent arms \\
\hline
squat & A person holds a barbell across the upper back, lowers the body by bending the knees and hips, and then rises back to a standing position \\
\hline
w raise & A person raises dumbbells in front of the body to form a ``W'' shape with the arms, lifting them to shoulder height \\
\hline
burpees & A person drops into a squat, places the hands on the floor, kicks the feet back into a plank, returns to a squat, and jumps upward \\
\hline
clean and press & A person lifts a barbell from the floor to the shoulders and then presses it overhead by extending the arms \\
\hline
drag curl & A person holds a barbell and curls it upward by keeping it close to the body, dragging it along the torso while bending the elbows \\
\hline
overhead extension thruster & A person holds a weight at shoulder height, performs a squat, and then extends the arms to press the weight overhead while rising to a standing position \\
\hline
standing ab twists & A person stands, lifts one knee, twists the torso to touch the opposite elbow to the raised knee, then alternates sides \\
\hline
warmup 2 & A person steps forward with the left foot into a lunge while raising both arms overhead, then returns to a standing position \\
\hline
warmup 4 & A person stands with feet shoulder-width apart and rotates the torso, reaching the right hand toward the left and back \\
\hline
warmup 9 & A person stands with feet shoulder-width apart and alternately reaches each arm upward \\
\hline
warmup 10 & A person stands upright, first bends the elbows and lifts them to shoulder height, then straightens the arms and stretches them out to the sides at shoulder level \\
\hline
warmup 13 & A person stands upright and stretches both arms straight up, then lowers them straight down to the ground \\
\hline
warmup 17 & A person stands upright and stretches both arms straight overhead \\
\hline
warmup 18 & A person squats down and then rises, extending both arms straight overhead \\
\hline
warmup 19 & A person bends into a squat, then explosively jumps upward while extending both arms straight overhead, fully lengthening the body \\
\hline

\end{longtable}
\endgroup

\begin{figure}[t] 
    \centering
    \includegraphics[width=\linewidth, trim=5mm 40mm 5mm 0mm, clip]{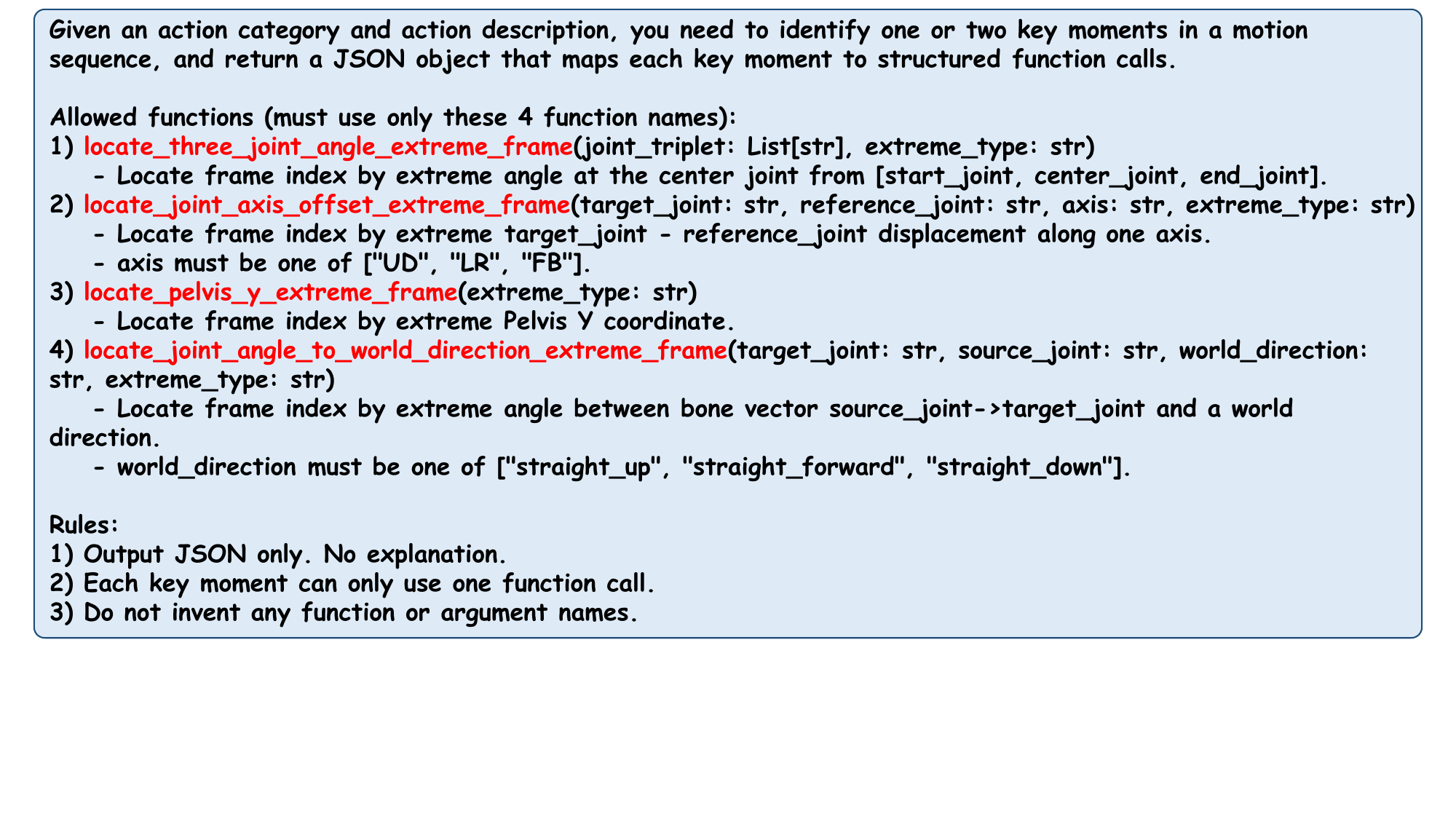} 
    \setlength{\fboxsep}{0.5pt}
    \vspace{-3mm}
    \caption{Prompt for frame-level queries.}
    \label{fig:frame_query_prompt} 
\end{figure}

\begin{figure}[t] 
    \centering
    \includegraphics[width=\linewidth, trim=5mm 45mm 5mm 0mm, clip]{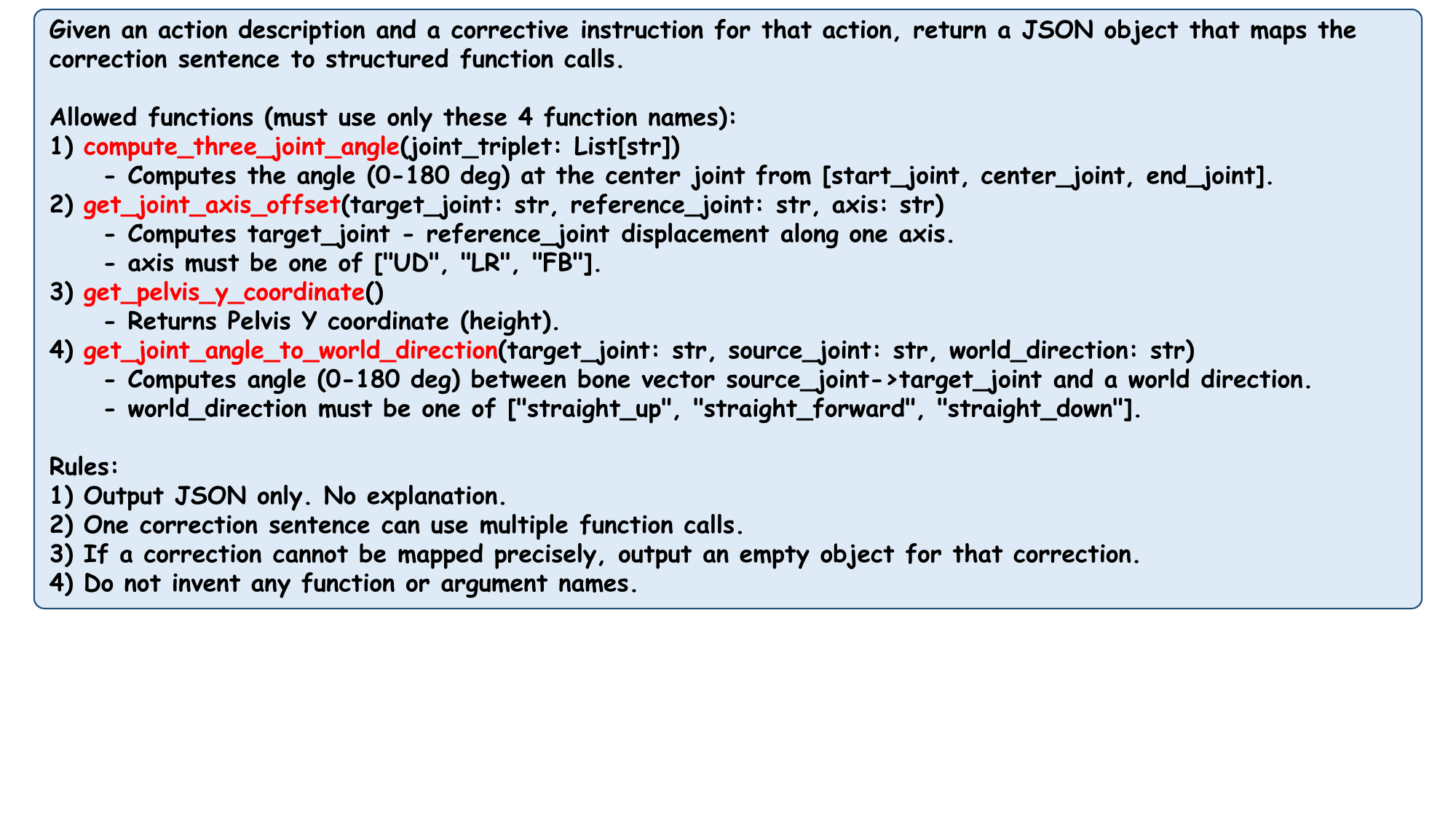} 
    \setlength{\fboxsep}{0.5pt}
    \vspace{-3mm}
    \caption{Prompt for kinematic queries.}
    \label{fig:kinematic_query_prompt} 
\end{figure}

\section{Semantic Parser in PPV}
\label{sec:appendix_parser}

\subsection{Parser Design and Implementation}

To operationalize the semantic parsing process described in the main text, we employ a powerful language model, Gemini-3-Flash, as the parser $\mathcal{P}$. This choice reflects a deliberate trade-off between economic cost and performance, as our goal is not to introduce a new model or achieve state-of-the-art results. Rather, PPV is designed as an enhanced baseline to study whether models can correctly reason over explicitly grounded kinematic measurements. In line with this objective, we focus on advancing the understanding of model behavior, rather than demonstrating that one model outperforms another. This is why we evaluate PPV across multiple models.

Specifically, PPV enables us to investigate whether providing structured, physically meaningful measurements can reduce hallucinations in motion reasoning. If models are able to make correct judgments when grounded on such measurements, this suggests that hallucinations may largely arise from ambiguity in the video modality, where models struggle to form a precise and global understanding of motion dynamics. From this perspective, PPV offers a promising direction for mitigating kinematic hallucinations by introducing explicit measurement-based grounding. The experiments in the main paper show that PPV leads to consistent improvements across all evaluated models, supporting the hypothesis that kinematic grounding can effectively reduce hallucinations in motion instruction tasks.

Given a motion instruction $i$ and action specification $\mathcal{A}$, the semantic parser $\mathcal{P}$ generates two queries, $Q_f$ and $Q_k$, as defined in the main text. Intuitively, $Q_f$ determines where to look in the motion sequence by localizing key frames based on kinematic extrema, while $Q_k$ specifies what to measure at those frames in terms of physical quantities. In practice, the parser also needs to establish the correspondence between key moments and the instruction, forming a complete and executable query. Together, they translate high-level language descriptions into executable kinematic queries that can be applied to motion data.

To construct these queries, we adopt a constrained prompting strategy (see \autoref{fig:frame_query_prompt} and \autoref{fig:kinematic_query_prompt}) that maps natural language instructions to structured function calls. The parser is restricted to a small set of atomic kinematic measurement functions, including joint angle computation, joint displacement along canonical axes, pelvis height estimation, and joint orientation with respect to world directions. These functions are intentionally designed to be minimal and general, allowing diverse motion instructions to be expressed through composition without introducing unnecessary complexity.

Importantly, the use of only four atomic functions avoids imposing handcrafted mappings between instructions and functions or between action categories and measurement types. This design reduces human bias and limits implicit assumptions in the parsing process, enabling the semantic parser to operate in a more flexible and generalizable manner across different motion contexts.

The prompt further enforces strict output constraints, requiring the parser to produce structured JSON outputs using only the predefined function set. Multiple function calls are allowed when necessary to capture complex instructions, while cases that cannot be precisely grounded are explicitly mapped to empty outputs. This design ensures that the resulting queries remain interpretable, consistent, and directly executable, facilitating reliable extraction of kinematic measurements for downstream reasoning.

\begin{table}[htbp]
\centering
\setlength{\tabcolsep}{4pt}
\caption{Key-frame localization on ground-truth motion. N denotes the number of evaluated frames. Mean and median errors are computed from the absolute difference between predicted and human-annotated frame indices. Errors are measured in frames. Recall@$k$ is the percentage of predictions within $k$ frames of the human-annotated frame.}
\label{tab:parser_localization}
\begin{tabular}{lrrrrr}
\toprule
\textbf{Phase subset} & \textbf{N} & \textbf{MAE} & \textbf{Median} & \textbf{Recall@10 (\%)} & \textbf{Recall@24 (\%)} \\
\midrule
All & 1,782 & 10.38 & 2 & 76.88 & 86.59 \\
Default phase & 588 & 2.46 & 0 & 94.56 & 94.90 \\
Other key phases & 1,194 & 14.28 & 5 & 68.17 & 82.50 \\
\bottomrule
\end{tabular}
\end{table}

\begin{table}[htbp]
\centering
\setlength{\tabcolsep}{4pt}
\caption{Automated semantic audit of 306 non-empty query sets. Values are percentages. Mean is the average pass rate across the two evaluator runs, and inter-run agreement measures consistency between their judgments.}
\label{tab:parser_semantic_validity}
\begin{tabular}{lrrrr}
\toprule
\textbf{Criterion} & \textbf{Run 1} & \textbf{Run 2} & \textbf{Mean} & \textbf{Inter-run agreement} \\
\midrule
Joint/laterality relevance & 99.67 & 99.35 & 99.51 & 99.02 \\
Measurement validity & 98.37 & 99.02 & 98.69 & 98.04 \\
\bottomrule
\end{tabular}
\end{table}

\subsection{Parser Evaluation}
\label{sec:parser_evaluation}

We evaluate key-frame localization and query semantics separately. For $Q_f$, we run the localization functions on ground-truth (GT) motion to exclude errors from 3D motion reconstruction. We compare the predicted frames with 1,782 human-annotated frames from 891 corrective instructions, with one annotated frame in each video of a pair. All frame errors are measured at 50 FPS. Table~\ref{tab:parser_localization} reports the mean and median absolute frame errors, together with the percentage of predictions within 10 and 24 frames of the annotated frame. The median error is 2 frames (0.04 s) overall and 5 frames (0.10 s) for other key phases. We report results separately for the default phase and the remaining key phases of each action.

For $Q_k$, we assess whether the generated queries select appropriate joints and measurements for each corrective instruction. Of 312 instructions, 306 produce non-empty query sets. An additional evaluator model assesses these sets using fixed evaluation criteria and a temperature of 0. It checks whether the selected joints match the relevant body part and left/right side, and whether the selected measurement function and axis or world direction can capture the described motion difference. Measurement validity is assessed in separate requests that evaluate only this criterion. We repeat the full evaluation twice. Table~\ref{tab:parser_semantic_validity} reports the pass rates from both runs, their mean, and the percentage of query sets receiving the same judgment in both runs. The mean pass rates are 99.51\% for body-part and side relevance and 98.69\% for measurement validity. Agreement between runs is 99.02\% and 98.04\%, respectively. The six instructions that produce empty query sets are counted as parser abstentions (6/312, 1.92\%) and excluded from the semantic evaluation.

\section{More Experiments and Analysis}

\subsection{Statistical Reliability and Action Sensitivity}
\label{sec:statistical_reliability}

Multiple QA items share the same underlying video pair, so we use pairs as clusters in 10,000 action-stratified bootstrap resamples. Pairs are sampled with replacement within each of the 32 actions, keeping their QA items together and using the same draws across models and Base/PPV-GT conditions. We report percentile 95\% confidence intervals for each condition and its paired gain, with overall accuracy computed as the mean of DH, AH, and TH accuracies as in \autoref{tab:main-results}. To assess sensitivity to action selection, we remove all QA items from one action category at a time and recompute overall accuracy and the PPV-GT gain on the remaining 31 categories. The leave-one-action-out (LOAO) gain range reports the minimum and maximum gains across these 32 evaluations.

As shown in \autoref{tab:bootstrap_loao}, the 95\% confidence intervals for PPV-GT gains are entirely above zero for all five models. Each model also retains a positive gain after any one action category is excluded. These results indicate that the observed overall gains are not driven by a single action category within the 32 evaluated Fit3D categories.

\begin{table}[h]
\centering
\caption{Overall accuracy under Base and PPV-GT. Accuracy is reported in percent; gains (PPV-GT minus Base) are in percentage points. Brackets denote 95\% bootstrap confidence intervals. Gains use unrounded scores. The LOAO range gives the minimum and maximum gains across 32 action exclusions.}
\label{tab:bootstrap_loao}
\resizebox{\linewidth}{!}{%
\begin{tabular}{lcccc}
\toprule
\textbf{Model} & \textbf{Base [95\% CI]} & \textbf{PPV-GT [95\% CI]} & \textbf{Gain [95\% CI]} & \textbf{LOAO gain range} \\
\midrule
InternVL3.5-8B & 54.95 [52.58, 57.39] & 61.53 [59.21, 63.90] & +6.58 [4.16, 9.08] & 6.06--7.21 \\
LLaVA-OV-1.5-8B & 48.37 [46.37, 50.43] & 59.65 [57.70, 61.59] & +11.28 [9.07, 13.51] & 10.76--12.26 \\
Qwen3-VL-8B & 61.31 [59.14, 63.53] & 66.27 [63.94, 68.60] & +4.96 [2.60, 7.29] & 4.28--5.28 \\
Gemini-3-Flash & 67.21 [65.21, 69.24] & 85.23 [83.24, 87.22] & +18.03 [15.30, 20.62] & 17.38--18.68 \\
Qwen3.5-plus & 71.28 [69.24, 73.27] & 83.03 [81.05, 85.05] & +11.75 [9.50, 14.03] & 10.97--12.36 \\
\bottomrule
\end{tabular}%
}
\end{table}

\subsection{Response Bias in the DH Task}
\label{sec:dh_response_bias}

Original and Reversed instances each comprise half of DH, with correct answers uniformly ``Yes'' and ``No'', respectively. Table~\ref{tab:dh_response_bias} compares Base with PPV using reconstructed motion.

For Gemini-3-Flash and Qwen3.5-plus, PPV moves the ``Yes'' response rate closer to 50\%, and their Reversed gains outweigh their Original decreases. However, this effect is not consistent across models: Qwen3-VL-8B becomes more inclined to answer ``Yes'', while LLaVA-OV-1.5-8B remains strongly affirmative-biased and shows a slight decline in DH average accuracy. Thus, although PPV improves overall benchmark performance, its ability to mitigate pre-existing response biases remains limited for some models.

\begin{table}[htbp]
\centering
\setlength{\tabcolsep}{4pt}
\caption{DH accuracy changes (PPV minus Base, in percentage points) and ``Yes'' response rates (\%) using reconstructed motion.}
\label{tab:dh_response_bias}
\begin{tabular}{lrrrrr}
\toprule
\multirow{2}{*}{\textbf{Model}} & \multirow{2}{*}{$\Delta$\textbf{Original}} & \multirow{2}{*}{$\Delta$\textbf{Reversed}} & \multirow{2}{*}{$\Delta$\textbf{DH Avg}} & \multicolumn{2}{c}{\textbf{``Yes'' rate}} \\
\cmidrule(lr){5-6}
 & & & & \textbf{Base} & \textbf{PPV} \\
\midrule
InternVL3.5-8B & $-1.28$ & $+3.85$ & $+1.28$ & 65.88 & 63.30 \\
LLaVA-OV-1.5-8B & $+0.97$ & $-1.92$ & $-0.49$ & 98.24 & 99.68 \\
Qwen3-VL-8B & $+10.58$ & $-2.88$ & $+3.85$ & 56.41 & 63.14 \\
Gemini-3-Flash & $-9.62$ & $+44.87$ & $+17.63$ & 76.39 & 49.15 \\
Qwen3.5-plus & $-3.10$ & $+41.56$ & $+19.23$ & 75.96 & 53.63 \\
\bottomrule
\end{tabular}
\end{table}

\subsection{Further Analysis on the AH Task}

We further decompose the AH task into two settings: same joint and different joints, based on whether the compared instructions refer to the same body part. The results are summarized in \autoref{tab:ah-results}. Both settings yield comparable performance across base models. In particular, smaller models achieve 59\% average accuracy, only marginally above the theoretical random baseline, indicating that the task provides an effective test of models' fine-grained motion reasoning ability.

This behavior indicates that the proposed setting successfully mitigates shortcut learning based on joint-level priors or linguistic patterns. Even when the same joint is involved, models struggle to capture subtle kinematic differences, while in the different-joint setting, they fail to leverage simple body-part distinctions as reliable signals. Together, these findings highlight that the difficulty of the task primarily arises from fine-grained motion reasoning rather than semantic ambiguity at the language level.

With the introduction of PPV, most models exhibit consistent performance improvements. This suggests that explicit kinematic grounding helps bridge the gap between language and motion, enabling models to make more reliable judgments. Notably, stronger closed-source models benefit more significantly, with performance gains reaching up to 8\%, likely due to their superior reasoning capabilities in integrating structured measurements. Overall, these results further support our hypothesis that hallucinations are closely tied to the lack of precise kinematic grounding, and that evidence-based reasoning provides a promising direction for mitigating such issues.

\begin{table}[t]
\centering
\caption{Additional results of the AH task. Models with PPV are shown in gray. Subscripts denote performance change relative to the base model (\textcolor{teal}{blue}: gain, \textcolor{red}{red}: loss). Values are accuracy (\%). Best in \textbf{bold}, second best \underline{underlined}.}
\label{tab:ah-results}

\newcommand{\gain}[1]{\textsubscript{\textcolor{teal}{\scriptsize($\uparrow$#1)}}}
\newcommand{\loss}[1]{\textsubscript{\textcolor{red}{\scriptsize($\downarrow$#1)}}}
\begin{tabular}{lccc}
\toprule
\multirow{2}{*}{\textbf{Model}} & \multicolumn{3}{c}{\textbf{AH}} \\ \cmidrule(lr){2-4}
 & Same Joint & Different Joints & Avg \\ \midrule

InternVL3.5-8B & 56.41 & 58.33 & 57.33 \\
\rowcolor{gray!20} +PPV & 66.66	& 57.98	& 62.5 \gain{5.17} \\ \addlinespace[0.5em]

LLaVA-OV-1.5-8B & 54.8 & 52.08 & 53.5 \\
\rowcolor{gray!20} +PPV & 59.93	& 53.81	& 57 \gain{3.5} \\ \addlinespace[0.5em]

Qwen3-VL-8B & 68.26	& 63.88	& 66.16 \\
\rowcolor{gray!20} +PPV & 69.23	& 58.33	& 64 \loss{2.16} \\ \midrule \addlinespace[0.5em]

Gemini-3-Flash & 84.93	& 81.94	& 83.5 \\
\rowcolor{gray!20} +PPV & 89.31	& 90.27	& \underline{89.77} \gain{6.27} \\ \addlinespace[0.5em]

Qwen3.5-plus & 82.26	& 81.13	& 81.72 \\ 
\rowcolor{gray!20} +PPV &  91.02	& 88.77	& \textbf{89.94} \gain{8.22} \\ 

\bottomrule
\end{tabular}%
\end{table}

\begin{table}[h]
\centering
\caption{Motion reconstruction error (MPJPE and PA-MPJPE)}
\label{tab:reconstruction_error}

\begin{tabular}{lcc}
\toprule
\textbf{Source} & \textbf{MPJPE} & \textbf{PA-MPJPE} \\
\midrule
Recon. (4D-Humans) & 330.8 & 38.8 \\
\bottomrule
\end{tabular}

\end{table}

\subsection{Impact of Motion Source}

Although we have reported in \autoref{sec:ablation} the performance gap of Qwen3.5-plus between reconstructed motion and ground-truth motion inputs, we further investigate this discrepancy across all evaluated models to better understand its implications and to explore the potential of kinematic measurement injection in real-world applications. Specifically, we compare model performance under two motion sources: ground-truth (GT) motion and reconstructed motion, aiming to assess how motion reconstruction artifacts affect downstream reasoning under the PPV framework.

As shown in \autoref{tab:reconstruction_error}, we report MPJPE and PA-MPJPE \cite{ionescu2013human3} between GT and reconstructed motions. The results indicate that the current motion reconstruction pipeline can faithfully recover fine-grained motion structures, as evidenced by relatively low PA-MPJPE. However, a noticeable gap in MPJPE suggests that there still exists a significant offset in global spatial position and scale, indicating that reconstructed motions may suffer from translation and proportion inconsistencies despite preserving local articulation patterns.

\autoref{tab:pred_gt_comparison} further presents the performance of five evaluated models (with PPV) across all tasks under GT and reconstructed motion inputs. The average performance gap remains within 2\%, suggesting that PPV is largely robust to global motion shifts introduced by reconstruction. This indicates that our kinematic measurement-based reasoning framework is primarily sensitive to structural and relational motion cues rather than absolute spatial alignment. However, these results are limited to the current Fit3D setting, as dedicated severe-occlusion stress tests and external validation in outdoor or other unconstrained scenarios have not been conducted. In these more challenging settings, occlusion-induced joint estimation errors may propagate to the extracted measurements and downstream judgments because PPV relies on reconstruction quality. This dependence also means that PPV can benefit from advances in motion reconstruction methods.

\begin{table}[t]
\centering
\caption{Performance under different motion sources (reconstructed vs. ground-truth motion) across DH, AH, TH tasks. Values are accuracy (\%).}
\label{tab:pred_gt_comparison}

\resizebox{\linewidth}{!}{
\begin{tabular}{lcccccccc}
\toprule
& \multicolumn{2}{c}{\textbf{DH}} 
& \multicolumn{2}{c}{\textbf{AH}} 
& \multicolumn{2}{c}{\textbf{TH}} 
& \multicolumn{2}{c}{\textbf{Avg}} \\
\cmidrule(lr){2-3} \cmidrule(lr){4-5} \cmidrule(lr){6-7} \cmidrule(lr){8-9}
\textbf{Model} 
& Recon. & GT 
& Recon. & GT 
& Recon. & GT 
& Recon. & GT \\
\midrule

InternVL3.5-8B   & 58.49 & 58.49 & 62.50 & 62.50 & 63.29 & 63.60 & 61.42 & 61.53 \\
LLaVA-OV-1.5-8B  & 49.67 & 49.83 & 57.00 & 56.33 & 74.36 & 72.78 & 60.34 & 59.65 \\
Qwen3-VL-8B      & 58.65 & 58.49 & 64.00 & 65.33 & 76.58 & 75.00 & 66.41 & 66.27 \\
Gemini-3-Flash   & 82.69 & 83.76 & 89.77 & 90.94 & 80.37 & 81.01 & 84.28 & 85.23 \\
Qwen3.5-plus     & 83.33 & 84.77 & 89.94 & 90.27 & 77.74 & 74.05 & 83.67 & 83.03 \\

\bottomrule
\end{tabular}
}
\end{table}


\end{document}